\pdfoutput=1
\documentclass[11pt]{article}
\usepackage[table,xcdraw]{xcolor}
\usepackage[preprint]{acl}
\usepackage{times}
\usepackage{hyperref}
\usepackage{latexsym}
\usepackage{booktabs}
\usepackage{multirow}
\usepackage{multicol}
\usepackage[T1]{fontenc}
\usepackage[utf8]{inputenc}
\usepackage{microtype}
\usepackage{inconsolata}
\usepackage{graphicx}
\usepackage{arydshln}
\usepackage{tcolorbox}
\usepackage{algorithm}
\usepackage{algorithmic}
\usepackage{amsmath, amssymb}
\tcbuselibrary{breakable}

\definecolor{lightgreen}{RGB}{0, 170, 0}

\title{DecIF: Improving Instruction-Following through Meta-Decomposition}

\author{
 \textbf{Tingfeng Hui\textsuperscript{1}},
 \textbf{Pengyu Zhu\textsuperscript{1}},
  \textbf{Bowen Ping\textsuperscript{2}}, \\
 \textbf{Ling Tang\textsuperscript{1}},
 \textbf{Guanting Dong\textsuperscript{1}},
 \textbf{Yaqi Zhang\textsuperscript{1}},
 \textbf{Sen Su\textsuperscript{1}\thanks{The corresponding author.}}
\\
\\
 \textsuperscript{1}Beijing University of Posts and Telecommunications, Beijing, China \\
 \textsuperscript{2}Peking University, Beijing, China \\
 \textsuperscript{1}\texttt{(huitingfeng,susen)@bupt.edu.cn}
}

\begin{document}
\maketitle
\begin{abstract}
Instruction-following has emerged as a crucial capability for large language models (LLMs). However, existing approaches often rely on pre-existing documents or external resources to synthesize instruction-following data, which limits their flexibility and generalizability. In this paper, we introduce DecIF, a fully autonomous, meta-decomposition guided framework that generates diverse and high-quality instruction-following data using only LLMs. DecIF is grounded in the principle of decomposition. For instruction generation, we guide LLMs to iteratively produce various types of meta-information, which are then combined with response constraints to form well-structured and semantically rich instructions. We further utilize LLMs to detect and resolve potential inconsistencies within the generated instructions. Regarding response generation, we decompose each instruction into atomic-level evaluation criteria, enabling rigorous validation and the elimination of inaccurate instruction-response pairs. Extensive experiments across a wide range of scenarios and settings demonstrate DecIF's superior performance on instruction-following tasks. Further analysis highlights its strong flexibility, scalability, and generalizability in automatically synthesizing high-quality instruction data. We release the source code and SFT data in \href{https://github.com/HypherX/DecIF}{Github} and \href{https://www.modelscope.cn/datasets/Hyphens/DecIF-10K-and-30K}{ModelScope}.
\end{abstract}

\begin{figure*}[htbp]
    \centering
    \resizebox{0.95\linewidth}{!}{%
        \includegraphics{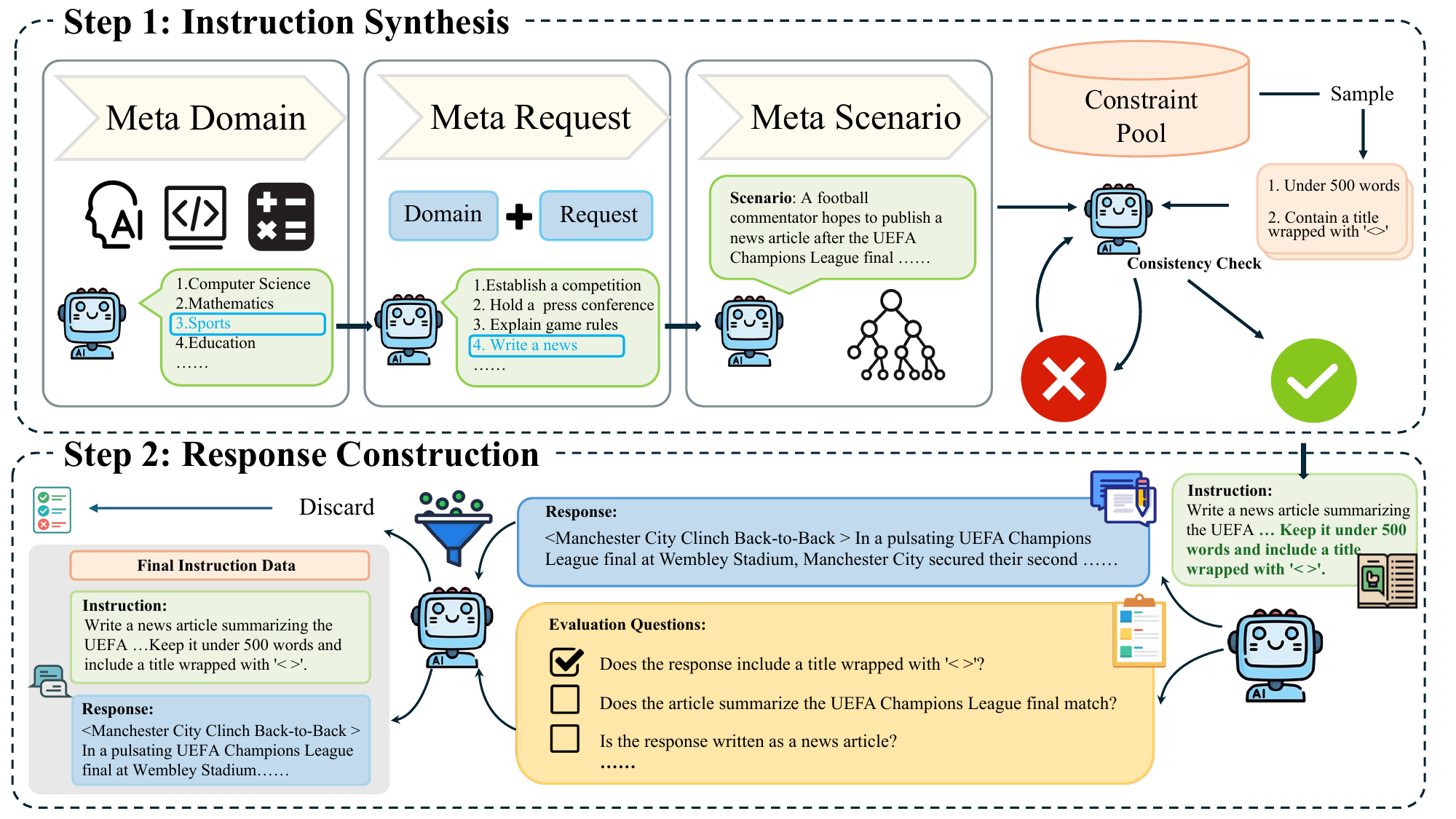}%
    }
    \caption{The overall workflow of DecIF.}
    \label{fig::method}
\end{figure*}

\section{Introduction}
Large language models (LLMs) have demonstrated strong performance across diverse NLP tasks and are increasingly integrated into daily life through chatbots and AI assistants such as Copilot \citep{openai2024gpt4technicalreport, yang2025qwen3technicalreport}. As their applications expand, instruction-following—the ability to accurately adhere to complex constraints—has become a key research focus \citep{li2024selfalignmentinstructionbacktranslation, dong2024selfplayexecutionfeedbackimproving, dong2024generalinstructionfollowingalignmentretrievalaugmented}. Improving this capability enables LLMs to reliably execute complex real-world instructions, advancing effective human-AI collaboration.

Recent efforts to align LLMs with instruction-following tasks have focused on two main directions. First, specialized optimization methods have been proposed to enhance instruction-following capabilities \citep{huang2025muscimprovingcomplexinstruction, sun2024coniferimprovingcomplexconstrained}. For example, \citet{zhang2024iopoempoweringllmscomplex} introduced IOPO, which incorporates both input and output preference pairs to improve alignment.
Second, significant work has focused on constructing high-quality instruction-following datasets \citep{wang2024instructionsconstraintslanguagemodel, liu2025aircomplexinstructiongeneration, he2024complexsimpleenhancingmulticonstraint}. WizardLM \citep{xu2023wizardlmempoweringlargelanguage} iteratively refines existing instructions to increase their complexity and diversity. AutoIF \citep{dong2024selfplayexecutionfeedbackimproving} integrates manually designed constraints into instructions and uses Python code to ensure consistency between inputs and outputs. AIR \citep{liu2025aircomplexinstructiongeneration} extracts instructions from documents and iteratively refines them to generate large-scale data. UltraIF \citep{an2025ultraifadvancinginstructionfollowing} synthesizes instruction data using a fine-tuned UltraComposer and employs LLM-as-a-judge for response verification.
However, most of these methods rely heavily on pre-existing documents or external resources, limiting their flexibility and diversity.

In this paper, we focus on addressing these limitations by proposing a novel framework, \textbf{DecIF}, a two-stage approach that constructs high-quality instruction-following data using only the internal capabilities of LLMs, without relying on any additional documents, external datasets, or human-annotated resources. Specifically,

(1) \textbf{Instruction Synthesis Stage}. Directly using LLMs to synthesize full instructions often results in unstable, low-quality or repetitive outputs due to the lack of structured guidance. Inspired by the step-by-step paradigm \citep{lightman2023letsverifystepstep, hsieh2023distillingstepbystepoutperforminglarger}, DecIF adopts a progressive approach by decomposing instruction generation into three distinct types of meta-information.
The process begins by prompting LLMs to generate meta-domains, which represent high-level conceptual categories such as \textit{Sports}, \textit{Technology} or \textit{Health}. These domains act as broad thematic boundaries, ensuring diversity and contextual relevance in the resulting instructions. Based on these meta-domains, we then prompt the model to generate meta-requests, which are general task formulations within each domain. For instance, under the \textit{Sports} domain, a meta-request might be \textit{Write sports news}. Meta-requests are intentionally abstract, avoiding specific scenarios or detailed constraints.
Building upon this two-step abstraction hierarchy, we proceed to generate meta-scenarios, which are concrete and context-rich situations derived from each meta-request. These scenarios are enriched with specific details such as personas, conditions, locations and time frames along with response constraints like style, format or length requirements. 
Combining the generated scenario with its associated constraints results in a fully-formed and semantically rich instruction.
By decomposing the instruction generation process into a sequence of structured subtasks, DecIF enables the systematic construction of rich and diverse instruction sets from scratch, leading to significant improvements in both quality and diversity.

(2) \textbf{Response Construction Stage}. DecIF leverages the principle of decomposition to enhance the precision and reliability of model responses. Specifically, the process begins by prompting LLMs to generate initial responses to the given instructions. The model is then guided to decompose each instruction into atomic-level evaluation criteria, ensuring that all requirements are identified and verifiable. Finally, these criteria are applied to rigorously evaluate and filter the generated responses, retaining only those that fully conform to the instruction for inclusion in the final dataset.

Extensive experiments across multiple models and comprehensive benchmarks demonstrate that DecIF significantly improves the instruction-following capabilities of LLMs compared to prior approaches, owing to the high quality of the synthesized data. Beyond basic instruction-following, we further evaluate our method across a range of common model capabilities, confirming the strong generalizability of DecIF-generated data.
To assess its practical applicability, we replace the instruction-following subset in Tulu-3 \citep{lambert2025tulu3pushingfrontiers} with our synthesized data. The resulting model surpasses Tulu-3 in instruction-following performance while remaining competitive across other capabilities, highlighting the compatibility of DecIF data with diverse training regimes.
We also investigate the potential of DecIF for generating general-purpose instruction data, as well as the impact of long-chain-of-thought (long-CoT) data on instruction-following performance. The main contributions of this paper are as follows:

(1) We introduce DecIF, a meta-decomposition guided framework for synthesizing high-quality instruction-following data without relying on pre-existing documents or external resources.

(2) DecIF decomposes instructions into three types of meta-information: domains, requests, and scenarios, which enable fine-grained control over diversity, complexity, and contextual richness, ensuring systematic and controllable generation.

(3) Extensive experiments show that DecIF outperforms existing methods in handling complex instructions, even surpassing the instruction-following subset in Tulu-3, and demonstrates strong potential for generating large-scale, general-purpose data compatible with open-source datasets.

\section{Methodology}

\subsection{Overview of DecIF}
DecIF generates high-quality instruction-following data through a two-stage process, without relying on any existing documents or datasets. As illustrated in Figure \ref{fig::method}, DecIF comprises two key stages: (1) \textbf{Instruction Synthesis Stage} and (2) \textbf{Response Construction Stage}, which are detailed in Sections \ref{sec::instruction_synthesis} and \ref{sec::response_construction}, respectively. Furthermore, in Section \ref{sec::discussion_of_decif}, we delve into the design philosophy of DecIF and offer in-depth reflections on its development. The complete procedure of DecIF is outlined in Algorithm \ref{alg::DecIF}. All prompt templates used in DecIF are provided in Appendix \ref{sec::prompt_template}.

\subsection{Instruction Synthesis Stage}
\label{sec::instruction_synthesis}
Previous efforts to construct instruction-following data have typically relied on real-world resources \citep{dong2024selfplayexecutionfeedbackimproving, an2025ultraifadvancinginstructionfollowing}, such as extracting authentic queries from ShareGPT \citep{vicuna2023} or leveraging PersonaHub \citep{ge2024scalingsyntheticdatacreation} to generate diverse instructions. In contrast, we aim to explore a fully from-scratch approach for constructing instruction-following data, enhancing the flexibility and adaptability of the method.

Recognizing that directly employing LLMs to synthesize complete instructions is often unstable and uncontrollable, frequently leading to numbers of low-quality and homogeneous data, we propose decomposing a complete instruction $i$ into three fine-grained components: \textit{domains}, \textit{requests}, and \textit{scenarios}. These components, collectively referred to as \textbf{meta-information}, form the foundation of our approach. We further divide the instruction synthesis process into three sub-tasks, ensuring greater precision and diversity in the generated data: $\mathcal{M}_d \xrightarrow{\text{generate}} \mathcal{M}_r \xrightarrow{\text{generate}} \mathcal{M}_s \xrightarrow{\text{synthesize}} i$ where $\mathcal{M}_d$, $\mathcal{M}_r$, and $\mathcal{M}_s$ denote the sets of meta-domains, requests, and scenarios.

\paragraph{Meta-Domains Generation} To ensure that the final instruction data remains within a reasonable and realistic scope, we leverage LLMs to generate meta-domains $\mathcal{M}_d = \{d_1, d_2, \dots, d_D\}$, which encompass diverse real-world interactions such as \textit{Artificial Intelligence} and \textit{Sports}. This process is straightforward: we prompt LLMs to generate a specified number $D$ of real-world domains in each iteration and subsequently filter out any duplicates.

\paragraph{Meta-Requests Generation} To ensure the instruction data is as comprehensive as possible, we prompt LLMs to generate diverse meta-requests $\mathcal{M}_r = \{r_1, r_2, \dots, r_R\}$ for each domain $d$. These requests capture the core needs to the domains without incorporating specific scenarios or contexts. For instance, in the domain of \textit{Artificial Intelligence}, the model might derive requests such as \textit{Develop a model} or \textit{Explain optimization methods}. By generating requests across various real-world domains, we establish an initial foundation for ensuring the diversity of the final instruction data.

\paragraph{Meta-Scenarios Generation} We then prompt LLMs to generate multiple scenarios $\mathcal{M}_s = \{s_1, s_2, \dots, s_S\}$ based on the diverse meta-requests, with each request serving as the core intent. These scenarios are enriched with specific details, including personas, conditions, locations, time, and other contextual elements. By leveraging these meta-scenarios, we further enhance the diversity of the instructions. Additionally, since the generated scenarios are progressively refined from real-world domains, this approach ensures that the resulting instructions remain both diverse and highly controllable.

Finally, following \citep{lambert2025tulu3pushingfrontiers}, we integrate the generated meta-scenarios $s \in \mathcal{M}_s$ and randomly sample several response constraints $\mathcal{C}_k \subseteq \mathcal{C}$ from a pre-defined constraint pool $\mathcal{C}$ according to a given probability distribution $\vec{p} = [p_1, p_2, p_3, p_4, p_5]$, where $p_k$ represents the probability of selecting exactly $k$ constraints from the pool. We then utilize the LLM $\theta$ to synthesize instructions $i$ conditioned on $s$ and $\mathcal{C}_k$. To further ensure the consistency of the instructions, we employ $\theta$ to detect conflicts among constraints and refine them using $\theta$ when necessary.
    
\begin{table*}[htbp]
\centering
\resizebox{\linewidth}{!}{%
\begin{tabular}{@{}lccccccccccc@{}}
\toprule
                                  &                                   & \multicolumn{4}{c}{\textbf{IFEval}} & \multicolumn{3}{c}{\textbf{Multi-IF}}                              & \multicolumn{2}{c}{\textbf{FollowBench}}    & \textbf{LiveBench}   \\ \cmidrule(l){3-12} 
\multirow{-2}{*}{\textbf{Method}} & \multirow{-2}{*}{\textbf{\#Data}} & Pr (S)  & In (S)  & Pr (L) & In (L) & Turn 1               & Turn 2               & Turn 3               & HSR                  & SSR                  & Score                \\ \midrule
\rowcolor[HTML]{EFEFEF} 
\multicolumn{12}{l}{\cellcolor[HTML]{EFEFEF}\textit{Directly utilize the open source dataset}}                                                                                                                                                        \\
ShareGPT                          & 10k                               & 39.00   & 50.24   & 42.70  & 53.96  & 40.25                     & 28.51                     & 23.13                     & 40.52                     & 57.97                     & 31.00                     \\
Evol-Instruct                     & 10k                               & 39.37   & 50.48   & 42.14  & 53.60  & 33.45                     & 21.36                     & 14.34                     & 29.78                     & 49.13                     & 26.90                     \\
Conifer                           & 13k                               & 44.36   & 56.24   & 48.98  & 60.55  & 44.61                     & 25.52                     & 17.95                     & 45.83                     & 59.98                     & 41.10                     \\
AIR                               & 10k                               & 43.99   & 55.16   & 47.32  & 58.51  & 42.00 & 23.56 & 18.94 & 44.97 & 61.54 & 38.50 \\
Condor                            & 20k                               & 55.64   & 64.99   & 58.60  & 67.15  & 50.74 & 30.73 & 23.05 & 48.10 & 61.92 & 41.10 \\  \midrule
\rowcolor[HTML]{EFEFEF} 
\multicolumn{12}{l}{\cellcolor[HTML]{EFEFEF}\textit{Utilize LLaMA-3.1-70B-Instruct as supervised model}} \\
AutoIF$^\dag$                            & 10k                               & 47.13   & 57.55   & 56.93  & 67.02  & 47.63                & 27.53                & 20.53                & -                    & 60.41                & 40.50                \\
UltraIF$^\dag$                           & 10k                               & 53.97   & 64.15   & 58.59  & 68.82  & 52.55                & 29.34                & 22.29                & -                    & 59.50                & 42.20                \\
DecIF                             & 10k                               & \textbf{70.98}   & \textbf{77.82}    & \textbf{73.75}   & \textbf{80.58}   & \textbf{64.75}                & \textbf{47.46}                & \textbf{35.98}                & \textbf{48.29}                & \textbf{62.28}                 & \textbf{50.50}                \\ \hdashline
\multicolumn{12}{l}{\cellcolor[HTML]{EFEFEF}\textit{Compare with more challenging dataset}} \\
Tulu-3-IF               & 30k                               & 68.58   & 77.22   & 72.27  & 80.58  & \textbf{70.14}                & 47.34                & 35.13                & 49.52                & 63.03                & 52.90                \\
UltraIF$^\ddag$ & 181k & 64.14 & 72.90 & 68.39 & 76.86 & 65.33 & 47.71 & \textbf{39.49} & \textbf{53.52}  & \textbf{67.20}  & 47.20 \\
DecIF                     & 30k                               & \textbf{71.72}    & \textbf{79.02}    & \textbf{74.86}   & \textbf{81.89}   & 69.57                & \textbf{51.27}                & 38.21                & 49.25                & 64.40                 & \textbf{54.30}                \\ \bottomrule
\end{tabular}
}
\caption{The main results of LLaMA-3.1-8B on four instruction-following benchmarks. Pr and In represent the prompt and instruction levels. S and L stand for strict and loose metrics for IFEval. For LiveBench, we only report the performance of instruction-following subset. Results marked with $\dag$ mean that we directly report the evaluation results in \citep{an2025ultraifadvancinginstructionfollowing}. $\ddag$ represents that we utilize the open-source 181k data from UltraIF.}
\label{tab::main_result}
\end{table*}
\begin{table}[htbp]
\centering
\resizebox{\linewidth}{!}{%
\begin{tabular}{@{}lccccc@{}}
\toprule
\multirow{2}{*}{\textbf{Method}} & \textbf{Code}      & \textbf{Reasoning} & \textbf{Math}  & \textbf{Conversation} & \textbf{General}             \\ \cmidrule(l){2-6} 
                                 & \textbf{HumanEval} & \textbf{BBH}       & \textbf{GSM8K} & \textbf{Arena Hard}   & \textbf{LiveBench {[}All{]}} \\ \midrule
AutoIF$^\dag$                           & 46.34              & 67.18              & 51.50          & 9.20                  & 17.50                        \\
UltraIF$^\dag$                          & 43.90              & 67.33              & 48.60          & 12.20                 & 21.30                        \\
DecIF                            & \textbf{46.95}     & \textbf{67.45}     & \textbf{60.42}  & \textbf{12.20}        & \textbf{21.90}     \\ \bottomrule      
\end{tabular}
}
\caption{The common capability results of LLaMA-3.1-8B on code, reasoning, math, conversation and general domains. Results marked with $\dag$ mean that we directly report the evaluation results in \citep{an2025ultraifadvancinginstructionfollowing}.}
\label{tab::general_result}
\end{table}

\subsection{Response Construction Stage}
\label{sec::response_construction}
To ensure accurate responses, we remain committed to the philosophy of decomposition. Specifically, we employ a two-stage response filtering strategy, termed \textbf{decompose then evaluate}, to eliminate inaccurate responses. The two stages are outlined as follows:

\paragraph{Instruction Decomposition} As shown in Figure \ref{fig::method}, given that each instruction $i$ often encompasses multiple requirements, directly using LLM $\theta$ to evaluate response correctness can lead to an inability to fully account for the context, resulting in partial inaccuracies. To address this challenge, we first prompt $\theta$ to decompose instructions into atomic-level components, breaking down all fine-grained requirements or constraints into individual evaluation criteria $Q = \{q_1, q_2, \dots, q_n\}$. For instance, for the instruction \textit{Write a four-line poem about Spring,} we can derive two evaluation questions: \textit{Is the response a four-line poem?} and \textit{Is the poem about Spring?} Using these atomic-level evaluation questions, we are able to assess responses accurately from various perspectives.

\begin{table*}[htbp]
\centering
\resizebox{\linewidth}{!}{%
\begin{tabular}{@{}lcllllcccccc@{}}
\toprule
                                  &                                   & \multicolumn{4}{c}{\textbf{IFEval}}       & \multicolumn{3}{c}{\textbf{Multi-IF}}      & \multicolumn{2}{c}{\textbf{FollowBench}}    & \textbf{LiveBench}   \\ \cmidrule(l){3-12} 
\multirow{-2}{*}{\textbf{Method}} & \multirow{-2}{*}{\textbf{\#Data}} & \multicolumn{1}{c}{Pr (S)}                   & \multicolumn{1}{c}{In (S)}                   & \multicolumn{1}{c}{Pr (L)}                   & \multicolumn{1}{c}{In (L)}                   & Turn 1               & Turn 2               & Turn 3               & HSR                  & SSR                  & Score                \\ \midrule
\rowcolor[HTML]{EFEFEF} 
\multicolumn{12}{l}{\cellcolor[HTML]{EFEFEF}\textit{Directly utilize the open source dataset}} \\
ShareGPT                          & 10k                               & 49.91                                        & 60.43                                        & 52.87                                        & 63.79                                        & 55.36                     & 39.56                     & 31.79                     & 51.02                     & 67.35                     & 38.60                     \\
Evol-Instruct                     & 10k                               & 56.01                                        & 66.91                                       & 60.44                                        & 70.74                                        & 59.40                     & 41.08                     & 31.23                     & 47.88                     & 66.37                     & 32.40                     \\
Conifer                           & 13k                               & 57.49                                        & 67.63                                        & 64.33                                        & 73.50                                        & 64.19                     & 22.59                     & 17.25                     & 63.05                     & 73.13                     & 48.60                     \\
AIR                               & 10k                               & 62.48                                        &  72.42                                       & 67.47                                        & 76.38                                        & 66.84 & 49.17 & 37.89 & 62.50 & 74.18 & 43.20 \\
Condor                            & 20k                               & 59.33                                        & 69.66                                       & 64.33                                        & 73.98                                        & 70.20 & 50.35 & 39.10 & 63.35 & 75.38 & 46.80 \\ \midrule
\multicolumn{12}{l}{\cellcolor[HTML]{EFEFEF}\textit{Utilize LLaMA-3.1-70B-Instruct as supervised model}} \\
DecIF                    & 10k                               & \textbf{76.89} & \textbf{83.21} & \textbf{78.56} & \textbf{85.13} & \textbf{80.76}                     & \textbf{64.59}                     & \textbf{52.98}                     & \textbf{64.84}                     & \textbf{76.95}                    & \textbf{55.00}                     \\ \hdashline
\multicolumn{12}{l}{\cellcolor[HTML]{EFEFEF}\textit{Compare with more challenging dataset}} \\
Tulu-3-IF                & 30k                               & 76.16                                        & 82.61                                        & 79.30                                        & 85.25                                        & \textbf{81.06}                     & 61.85                     & 47.12                     & \textbf{69.00}                     & \textbf{78.51}                     & 41.30                     \\
UltraIF$\ddag$ & 181k & 68.02 & 76.98 & 71.53 & 79.62 & 72.26 & 54.75 & 45.20 & 62.12 & 73.76 & 49.60 \\
DecIF                     & 30k                               & \textbf{78.00} & \textbf{83.57} & \textbf{79.48} & \textbf{85.61} & 80.98                    & \textbf{66.14}                     & \textbf{53.16}                     & 67.83                     & 77.57                   & \textbf{53.70}                     \\
\bottomrule
\end{tabular}
}
\caption{The main results of Qwen-3-8B-Base on four instruction-following benchmarks. Pr and In represent the prompt and instruction levels. S and L stand for strict and loose metrics for IFEval. For LiveBench, we only report the performance of instruction-following subset. $\ddag$ represents that we utilize the open-source 181k data from UltraIF.}
\label{tab::main_result_qwen}
\end{table*}
\begin{table}[htbp]
\centering
\resizebox{\linewidth}{!}{%
\begin{tabular}{@{}lccccc@{}}
\toprule
\textbf{Method}       & \textbf{\#Data} & \textbf{IFEval} & \textbf{Multi-IF} & \textbf{FollowBench} & \textbf{LiveBench} \\ \midrule
w/o judge & 10k             & 71.30                & 48.66                  & 52.97                     & 48.20                   \\
w/o filter & 10k             & 72.14                & 48.72                  & 53.68                     & 49.90                   \\
DecIF                 & 10k             & \textbf{75.78}  & \textbf{49.40}    & \textbf{55.29}       & \textbf{50.50}     \\ \hdashline
w/o judge & 30k             & 73.37                & 51.36                  & 53.77                     & 52.10                   \\
w/o filter & 30k             & 73.92                & 51.01                  & 54.26                     & 53.20                   \\
DecIF                 & 30k             & \textbf{76.87}  & \textbf{53.02}    & \textbf{56.83}       & \textbf{54.30}     \\ \bottomrule
\end{tabular}
}
\caption{The ablation study of instruction consistency judgement and response filtering. 'w/o judge' means that we do not perform instruction consistency judgement. 'w/o filter' represents training directly on the original instruction dataset. We report the average performance for each benchmark of LLaMA-3.1-8B.}
\label{tab::ablation_study}
\end{table}

\paragraph{Response Filtering} Next, we prompt $\theta$ to perform fine-grained evaluations of the responses $y$ based on the evaluation criteria $Q$, generating an evaluation list $\mathcal{E} = \{e_i\}_{i=1}^n$, where each $e_i \in \{\text{YES}, \text{NO}\}$ indicates whether the response satisfies the corresponding requirement. We retain only those samples $(i, y)$ for which all $e_i = \text{YES}$, discarding the rest and thereby constructing the final instruction dataset $\mathcal{D}$.

\subsection{Discussion of DecIF}
\label{sec::discussion_of_decif}
In this section, we present the design philosophy and training strategy of DecIF, a fully self-contained framework for synthesizing high-quality instruction-following data using only LLMs, without relying on external resources. Unlike methods that incorporate complex training strategies such as iterative DPO \citep{dong2024selfplayexecutionfeedbackimproving, an2025ultraifadvancinginstructionfollowing}, we focus on generating strong SFT data in a straightforward manner similar to \citep{lambert2025tulu3pushingfrontiers}.
We argue that high-quality SFT data forms a solid foundation for instruction-following capabilities and offers greater flexibility across model architectures and downstream tasks. 
For example, LLaMA-3 \citep{grattafiori2024llama3herdmodels} highlight the importance of well-curated SFT data in post-training stages. DecIF aims to enhance the quality of instruction-response pairs from scratch, without introducing complex training pipelines. In this paper, we only apply standard SFT to base models and compare the resulting performance with existing baselines.

\section{Experiments}

\subsection{Experimental Setup}
\textbf{Baselines.} Following \citep{an2025ultraifadvancinginstructionfollowing}, we employ LLaMA-3.1-70B-Instruct as the supervised model for the main results. Additionally, we explore alternative settings, such as Qwen-3-32B and GPT-4o, to assess the robustness of DecIF. 
In our experiments, we compare DecIF with a wide range of baselines, including ShareGPT \citep{vicuna2023}, Evol-Instruct \citep{xu2023wizardlmempoweringlargelanguage}, Conifer \citep{sun2024coniferimprovingcomplexconstrained}, Condor \citep{cao2025condorenhancellmalignment}, AIR \citep{liu2025aircomplexinstructiongeneration}, AutoIF \citep{dong2024selfplayexecutionfeedbackimproving}, and UltraIF \citep{an2025ultraifadvancinginstructionfollowing}. Furthermore, we incorporate the instruction-following subset of Tulu-3 \citep{lambert2025tulu3pushingfrontiers} and the open-source 181k data from UltraIF as more challenging baselines.

\textbf{Evaluation Benchmarks.} We evaluate our approach using four instruction-following benchmarks: IFEval \citep{zhou2023instructionfollowingevaluationlargelanguage}, Multi-IF \citep{he2024multiifbenchmarkingllmsmultiturn}, FollowBench \citep{jiang2024followbenchmultilevelfinegrainedconstraints}, and LiveBench \citep{white2025livebenchchallengingcontaminationlimitedllm}. 
For common capabilities, we employ a variety of specialized benchmarks: GSM8K \citep{cobbe2021trainingverifierssolvemath} and MATH \citep{hendrycks2021measuringmathematicalproblemsolving} for mathematical reasoning; HumanEval \citep{chen2021evaluatinglargelanguagemodels} for code generation; BBH \citep{suzgun2022challengingbigbenchtaskschainofthought} and HellaSwag \citep{zellers2019hellaswagmachinereallyfinish} for reasoning; GPQA \citep{rein2023gpqagraduatelevelgoogleproofqa} and MMLU \citep{hendrycks2021measuringmassivemultitasklanguage} for knowledge assessment; Arena Hard \citep{li2024crowdsourceddatahighqualitybenchmarks} for conversational abilities; and LiveBench (all) \citep{white2025livebenchchallengingcontaminationlimitedllm} for general capabilities.

For detailed information about the baselines, evaluation benchmarks, and training details, please refer to Appendix \ref{sec::baseline}, \ref{sec::benchmark}, \ref{sec::inplementation}, and \ref{sec::training}.

\begin{table*}[htbp]
\centering
\resizebox{\linewidth}{!}{%
\begin{tabular}{@{}lccccccccccc@{}}
\toprule
                                  &                                   & \multicolumn{4}{c}{\textbf{IFEval}} & \multicolumn{3}{c}{\textbf{Multi-IF}} & \multicolumn{2}{c}{\textbf{FollowBench}} & \textbf{LiveBench} \\ \cmidrule(l){3-12} 
\multirow{-2}{*}{\textbf{Method}} & \multirow{-2}{*}{\textbf{\#Data}} & Pr (S)  & In (S)  & Pr (L) & In (L) & Turn 1      & Turn 2     & Turn 3     & HSR                 & SSR                & Score              \\ \midrule
\multicolumn{12}{l}{\cellcolor[HTML]{EFEFEF}\textit{Results of LLaMA-3.1-8B}} \\ 
Tulu-3-IF               & 30k                               & 68.58   & 77.22   & 72.27  & 80.58  & 70.14       & 47.34      & 35.13      & 49.52               & 63.03              & 52.90              \\
DecIF (Qwen-3)                            & 30k                               & \textbf{69.87}   & \textbf{78.18}   & \textbf{74.12}  & 81.29  & \textbf{71.60}            & 51.39           & 34.61           & \textbf{60.99}                    & \textbf{71.32}                   & \textbf{55.00}                   \\ 
DecIF (GPT-4o)                            & 30k                               & 69.73   & 77.87   & 73.58  & \textbf{82.01}  & 70.17       & \textbf{51.50}      & \textbf{39.13}      & 50.38               & 65.10              & 54.40  \\ \midrule
\multicolumn{12}{l}{\cellcolor[HTML]{EFEFEF}\textit{Results of Qwen-3-8B-Base}}                      \\
Tulu-3-IF                & 30k                               & 76.16                                        & 82.61                                        & 79.30                                        & 85.25                                        & \textbf{81.06}                     & 61.85                     & 47.12                     & 69.00                     & 78.51                     & 41.30                     \\
DecIF (Qwen-3)                            & 30k                               & \textbf{78.56} & 84.65 & \textbf{81.89} & \textbf{87.53}  & 80.86            & \textbf{66.39}           & \textbf{55.32}           & \textbf{72.95}                   & \textbf{80.24}                   & \textbf{61.20}                   \\
DecIF (GPT-4o)                             & 30k                               & 78.37   & \textbf{85.01}   & 81.15  & 87.05  & 80.02       & 62.99      & 51.72      & 70.84               & 80.07              & 56.80  \\ \bottomrule         
\end{tabular}
}
\caption{The results on four instruction-following benchmarks with LLaMA-3.1-8B and Qwen-3-8B-Base. Qwen-3 and GPT-4o represent that we utilize Qwen-3-32B and GPT-4o as supervised models.}
\label{tab::more_supervised_model}
\end{table*}

\subsection{Main Results}
Table \ref{tab::main_result} presents the performance of DecIF on four instruction-following benchmarks compared to other baselines. We apply only SFT to base models, and the results demonstrate that DecIF significantly outperforms all baselines. Specifically, compared to UltraIF, our method achieves nearly 8-18\% improvements on IFEval, MultiIF, and LiveBench with the same amount of training data, along with nearly a 3\% enhancement on FollowBench. 
When scaling up to 30k training data, DecIF continues to achieve the best performance on most benchmarks, even when compared to the instruction-following subset of Tulu-3 and UltraIF-181k. Overall, DecIF does not rely on external resources, constructs instruction-following data entirely from scratch, and achieves substantial performance improvements over prior approaches and challenging datasets. This establishes DecIF as a more \textit{flexible and scalable framework for enhancing instruction-following capabilities}.

\begin{figure*}[htbp]
    \centering
    \includegraphics[width=\linewidth]{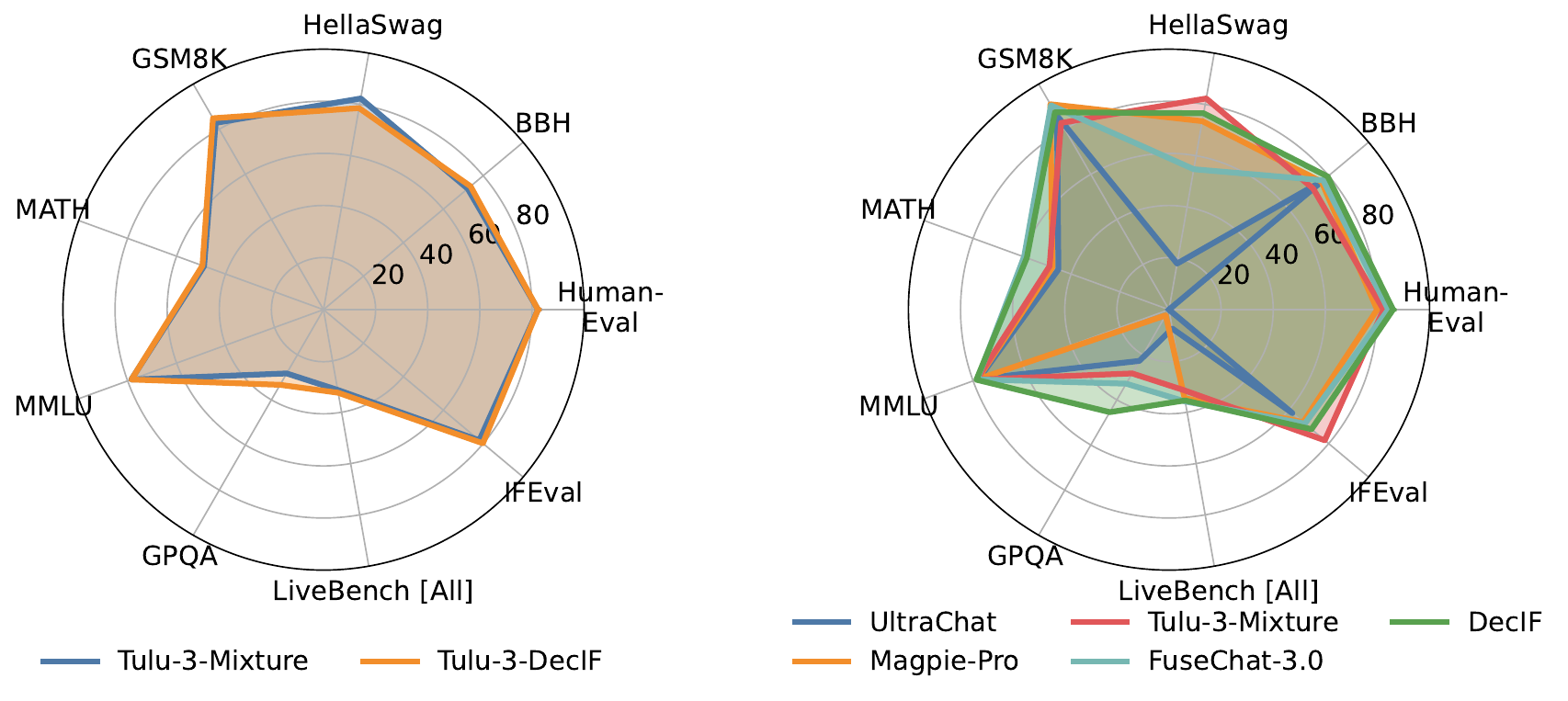}
    \caption{The results on Tulu-3-Mixture and Tulu-3-DecIF (Left) and the results of different large-scale general-purpose instruction data (Right). All experiments are conducted on Qwen-3-8B-Base.}
    \label{fig::mixture}
\end{figure*}

\subsection{Common Capabilities Evaluation}
To ensure that the constructed instruction-following data does not negatively impact or conflict with common capabilities, we follow \citep{an2025ultraifadvancinginstructionfollowing} and conduct a comprehensive evaluation across coding, math, reasoning, conversation, and general capabilities. Based on the data construction process of DecIF outlined earlier, the resulting instruction data is more diverse than previous methods, encompassing a broader range of instructions across various domains. 
As shown in Table \ref{tab::general_result}, DecIF outperforms AutoIF and UltraIF in terms of common capabilities, particularly excelling in math and coding domains. This further indicates that the instructions generated by DecIF are not only more diverse but also less likely to cause conflicts when integrated with data from other domains during training (detailed experiments and discussions on data mixing are provided in Section \ref{sec::mixture_with_other_training_data}).

\subsection{Results on Other Base Models}
To further validate the generalizability of the data we construct, we utilize the data presented in Table \ref{tab::main_result} to train Qwen-3-8B-Base. The performance results are summarized in Table \ref{tab::main_result_qwen}. We find that DecIF achieves superior performance on several benchmarks, notably outperforming the results obtained from Tulu-3's 30k and UltraIF's 181k training data while utilizing only 10k instruction data on most instruction-following benchmarks. This result highlights the versatility and robustness of DecIF. Note that UltraIF-181K performs less impressively on Qwen-3-8B-Base compared to its performance on LLaMA-3.1-8B. In contrast, DecIF demonstrates even stronger capabilities on the more advanced Qwen-3 model, suggesting its broader applicability, even on state-of-the-art models. Appendix \ref{sec::further_experiment} shows more results on various advanced models.

\subsection{Ablation Study}
To further validate the effectiveness of our decomposition-based response evaluation strategy, we directly compare the results with those obtained using unfiltered raw instruction data. As demonstrated in Table \ref{tab::ablation_study}, the filtered data consistently outperforms the unfiltered data across four instruction-following benchmarks for both the 10k and 30k datasets. This highlights the essential role of precise, high-quality data in achieving superior instruction-following performance.

\section{Further Analysis}

\subsection{Explore More Supervised Models}
To further explore the flexibility of DecIF, we utilize Qwen-3-32B and GPT-4o as supervised models to generate instruction-following data, which is subsequently used to train LLaMA-3.1-8B and Qwen-3-8B-Base. The detailed results are summarized in Table \ref{tab::more_supervised_model}. We find that using Qwen-3 and GPT-4o as supervised models consistently yields superior performance than Tulu-3, further underscoring the robustness of the DecIF method. This demonstrates that DecIF can effectively harness any LLMs to synthesize high-quality data.

\begin{figure}[htbp]
    \centering
    \includegraphics[width=\linewidth]{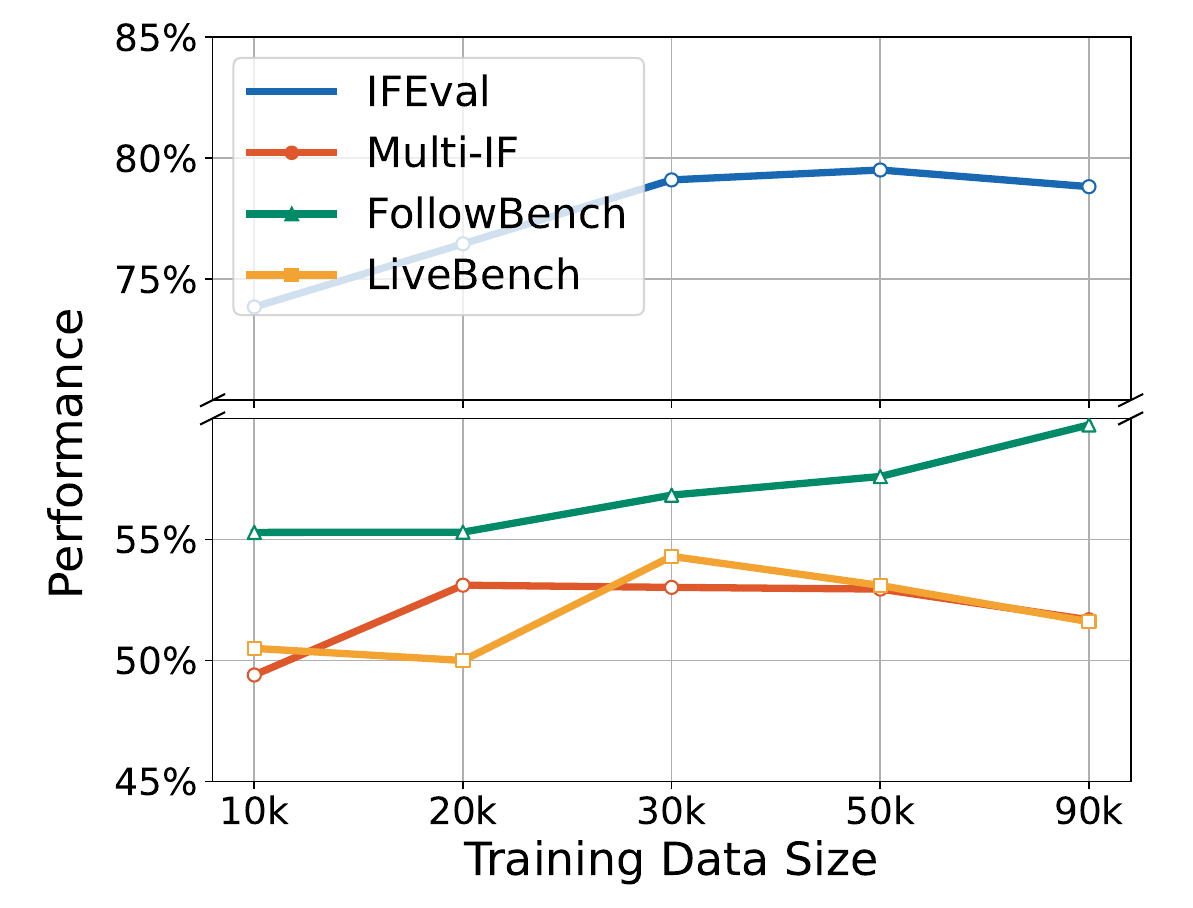}
    \caption{The results on four instruction-following benchmarks across different sizes of training data.}
    \label{fig::scaling}
\end{figure}

\subsection{Mixture with Other Training Data}
\label{sec::mixture_with_other_training_data}
Given that the SFT stage of recent advanced models \citep{grattafiori2024llama3herdmodels} typically relies on a substantial volume of high-quality, cross-domain data, we argue that top-tier instruction-following datasets should not only excel when trained in isolation, but also demonstrate strong compatibility with training data from diverse domains. Crucially, such datasets should avoid introducing conflicts that could undermine overall model performance.
To evaluate this, we replace the instruction-following component of Tulu-3 with data synthesized by DecIF. As shown in Figure \ref{fig::mixture} (Left), substituting the original data with our synthesized dataset results in improved performance on IFEval, while maintaining stable performance across other domains without any signs of degradation. This outcome clearly underscores the compatibility of our synthesized data with multi-domain training data, thereby further highlighting the broad applicability and potential of DecIF in diverse training contexts.

\subsection{Scaling Capability of DecIF}
In this section, we examine the scaling behavior of DecIF. Specifically, we employ LLaMA-3.1-70B-Instruct to generate instruction-following datasets ranging in size from 10k to 90k examples. As illustrated in Figure \ref{fig::scaling}, performance on IFEval, LiveBench, and Multi-IF initially improves with increasing dataset size but eventually plateaus or declines. We hypothesize that this trend stems from the risk of overfitting when rule-based evaluation metrics are used to assess precise instruction-following capabilities, leading to diminished returns as data size grows.
In contrast, FollowBench exhibits a consistently upward trajectory, with performance improving steadily as more data is introduced. This suggests that for more subjective aspects of instruction following, evaluated via GPT-4o, larger datasets offer greater diversity of scenarios, thereby supporting continued performance gains within the tested range. Overall, the model achieves its optimal balance across all metrics at a dataset size of 30k, which curiously aligns with the size of the instruction-following subset in Tulu-3.

\subsection{Large-Scale Data Synthesis of DecIF}
In the data construction process of DecIF, we observe its strong potential to synthesize large-scale, general-purpose instruction-following data. To explore this capability, we remove response constraints from the original pipeline and generate instruction data directly from a diverse set of meta-scenarios. Using Qwen-3-32B, we obtain approximately 150k high-quality instruction-response pairs after response filtering.
As shown in Figure \ref{fig::mixture} (Right), the large-scale general-purpose instruction data synthesized by DecIF achieves competitive performance when compared to recent advanced datasets such as UltraChat \citep{ding2023enhancingchatlanguagemodels}, Magpie-pro \citep{xu2024magpie}, Tulu-3-Mixture, and FuseChat-3.0 \citep{yang2025fusechat30preferenceoptimizationmeets}. Compared to UltraChat, Tulu-3 and FuseChat-3.0, DecIF does not rely on any pre-existing documents or external resources to synthesize high-quality instruction data. In contrast to Magpie, which requires complex filtering mechanisms to eliminate low-quality or unparseable outputs, DecIF employs a more interpretable approach based on decomposed instructions to filter responses. We believe that DecIF also holds great potential in generating instruction data. By incorporating additional control signals, such as instruction difficulty, the synthesis process can be further refined. This represents a promising direction for future research. Appendix \ref{sec::further_experiment} shows more exploration on recent long-CoT training.

\section{Related Work}
\textbf{Instruction-Following.} Instruction-following has become a critical capability for LLMs, enabling them to better understand and execute complex human instructions \citep{li2024selfalignmentinstructionbacktranslation, dong2024selfplayexecutionfeedbackimproving, dong2024generalinstructionfollowingalignmentretrievalaugmented}. Early efforts such as ShareGPT \citep{vicuna2023} relied on user-shared dialogues for model fine-tuning. To reduce reliance on costly human-labeled data, recent studies focus on synthetic data generation.
Instruction back-translation \citep{li2024selfalignmentinstructionbacktranslation} generates new instruction–response pairs from model outputs. Conifer \citep{sun2024coniferimprovingcomplexconstrained} improves complex instruction quality through iterative refinement. AutoIF \citep{dong2024selfplayexecutionfeedbackimproving} incorporates code execution feedback to validate and refine responses. UltraIF \citep{an2025ultraifadvancinginstructionfollowing} decomposes prompts into sub-queries and constraints, generating diverse and structured instructions. AIR \citep{liu2025aircomplexinstructiongeneration} enhances alignment by iteratively refining annotations, instructions, and responses.
In addition to data construction, specialized training strategies have been proposed to improve instruction-following. MuSc \citep{huang2025muscimprovingcomplexinstruction} introduces multi-granularity self-contrastive training with constraint-aware preference data and token-level supervision. IOPO \citep{zhang2024iopoempoweringllmscomplex} jointly optimizes instruction and response preferences within a unified framework, improving performance on complex tasks. These methods collectively enhance the alignment of LLMs with intricate instructions.

\textbf{Data Synthesis.} Data synthesis plays a key role in addressing the scarcity and high cost of manually annotated datasets, enabling the automatic generation of large-scale datasets for LLMs training. Self-Instruct \citep{wang-etal-2023-self-instruct} uses minimal seed prompts to bootstrap large-scale instruction sets. WizardLM \citep{xu2023wizardlmempoweringlargelanguage} iteratively evolves simple prompts into complex, multi-step tasks. Condor \citep{cao2025condorenhancellmalignment} combines knowledge-driven generation with self-refinement to improve dataset accuracy. PersonaHub \citep{ge2024scalingsyntheticdatacreation} synthesizes data across diverse persona-driven perspectives, capturing richer interaction patterns. Magpie \citep{xu2024magpie} fully automates instruction generation, eliminating the need for manual seeds and enhancing scalability.
Building on these advances, our method DecIF introduces a fully automated, two-stage framework for generating diverse and high-quality instruction-following datasets from scratch. Unlike existing approaches, DecIF does not rely on external resources or specialized training techniques, offering a flexible and scalable solution through a streamlined workflow.

\section{Conclusion}
In this paper, we propose DecIF, a novel, flexible, scalable, and generalizable approach for synthesizing high-quality instruction-following data, eliminating the need for any external documents or datasets. By decomposing the process of synthesizing complete instructions into step-by-step generation of meta domains, requests, and scenarios, we ensure the diversity of the generated instructions. Furthermore, adhering to the same philosophy of decomposition, we break down instructions into atomic-level evaluation criteria, enabling effective filtering of responses to obtain accurate instruction-response pairs. Extensive experiments demonstrate that DecIF not only achieves superior performance across various instruction-following benchmarks but also exhibits strong potential for generating general-purpose instruction data. Overall, DecIF provides a promising pathway for stably and efficiently synthesizing high-quality instruction data, paving the way for future advancements in the field.


\section*{Limitations}
In this paper, we propose DecIF, a method that enables the synthesis of high-quality instruction-following data from scratch, without relying on any existing documents or datasets. Despite DecIF's superior performance, it still has several limitations.

(1) Although we have explored the potential of DecIF in synthesizing large-scale, general-purpose instruction data, due to space and scope limitations, we do not further investigate its full capacity. In future work, we plan to conduct a more in-depth study of DecIF's ability to generate large-scale instruction data, with finer-grained control, to further advance the field.

(2) For the response validation process, we acknowledge that although DecIF is capable of filtering out erroneous responses to obtain accurate instruction-response pairs, this approach may introduce certain risks. Specifically, the model might benefit from retaining more challenging data to further enhance its capabilities. The current response filtering mechanism in DecIF could potentially lead to a bias in the final instruction data, favoring simpler instructions over more complex ones. In future work, we will continue to explore and improve the judgment and refinement methods for responses, aiming to preserve a broader range of data while ensuring response accuracy.

(3) Since DecIF relies entirely on LLMs for data synthesis, it exhibits a strong dependence on the capabilities of the underlying models. In future work, we will explore more effective control mechanisms to enable even smaller LLMs, such as those at the 8B scale, to generate high-quality instruction data.

(4) In this paper, we primarily focus on single-turn English instruction-following data. We plan to explore DecIF's potential in generating multi-turn dialogue data and instruction data in other languages in future work.

\bibliography{custom}

\begin{thebibliography}{40}
\providecommand{\natexlab}[1]{#1}

\bibitem[{An et~al.(2025)An, Sheng, Cui, Si, Ding, Cheng, and Chang}]{an2025ultraifadvancinginstructionfollowing}
Kaikai An, Li~Sheng, Ganqu Cui, Shuzheng Si, Ning Ding, Yu~Cheng, and Baobao Chang. 2025.
\newblock \href {https://arxiv.org/abs/2502.04153} {Ultraif: Advancing instruction following from the wild}.
\newblock \emph{Preprint}, arXiv:2502.04153.

\bibitem[{Cao et~al.(2025)Cao, Zhang, Li, Zhang, Liu, Duan, Zhang, and Chen}]{cao2025condorenhancellmalignment}
Maosong Cao, Taolin Zhang, Mo~Li, Chuyu Zhang, Yunxin Liu, Haodong Duan, Songyang Zhang, and Kai Chen. 2025.
\newblock \href {https://arxiv.org/abs/2501.12273} {Condor: Enhance llm alignment with knowledge-driven data synthesis and refinement}.
\newblock \emph{Preprint}, arXiv:2501.12273.

\bibitem[{Chen et~al.(2021)Chen, Tworek, Jun, Yuan, de~Oliveira~Pinto, Kaplan, Edwards, Burda, Joseph, Brockman, Ray, Puri, Krueger, Petrov, Khlaaf, Sastry, Mishkin, Chan, Gray, Ryder, Pavlov, Power, Kaiser, Bavarian, Winter, Tillet, Such, Cummings, Plappert, Chantzis, Barnes, Herbert-Voss, Guss, Nichol, Paino, Tezak, Tang, Babuschkin, Balaji, Jain, Saunders, Hesse, Carr, Leike, Achiam, Misra, Morikawa, Radford, Knight, Brundage, Murati, Mayer, Welinder, McGrew, Amodei, McCandlish, Sutskever, and Zaremba}]{chen2021evaluatinglargelanguagemodels}
Mark Chen, Jerry Tworek, Heewoo Jun, Qiming Yuan, Henrique~Ponde de~Oliveira~Pinto, Jared Kaplan, Harri Edwards, Yuri Burda, Nicholas Joseph, Greg Brockman, Alex Ray, Raul Puri, Gretchen Krueger, Michael Petrov, Heidy Khlaaf, Girish Sastry, Pamela Mishkin, Brooke Chan, Scott Gray, and 39 others. 2021.
\newblock \href {https://arxiv.org/abs/2107.03374} {Evaluating large language models trained on code}.
\newblock \emph{Preprint}, arXiv:2107.03374.

\bibitem[{Chiang et~al.(2023)Chiang, Li, Lin, Sheng, Wu, Zhang, Zheng, Zhuang, Zhuang, Gonzalez, Stoica, and Xing}]{vicuna2023}
Wei-Lin Chiang, Zhuohan Li, Zi~Lin, Ying Sheng, Zhanghao Wu, Hao Zhang, Lianmin Zheng, Siyuan Zhuang, Yonghao Zhuang, Joseph~E. Gonzalez, Ion Stoica, and Eric~P. Xing. 2023.
\newblock \href {https://lmsys.org/blog/2023-03-30-vicuna/} {Vicuna: An open-source chatbot impressing gpt-4 with 90\%* chatgpt quality}.

\bibitem[{Cobbe et~al.(2021)Cobbe, Kosaraju, Bavarian, Chen, Jun, Kaiser, Plappert, Tworek, Hilton, Nakano, Hesse, and Schulman}]{cobbe2021trainingverifierssolvemath}
Karl Cobbe, Vineet Kosaraju, Mohammad Bavarian, Mark Chen, Heewoo Jun, Lukasz Kaiser, Matthias Plappert, Jerry Tworek, Jacob Hilton, Reiichiro Nakano, Christopher Hesse, and John Schulman. 2021.
\newblock \href {https://arxiv.org/abs/2110.14168} {Training verifiers to solve math word problems}.
\newblock \emph{Preprint}, arXiv:2110.14168.

\bibitem[{Contributors(2023)}]{2023opencompass}
OpenCompass Contributors. 2023.
\newblock Opencompass: A universal evaluation platform for foundation models.
\newblock \url{https://github.com/open-compass/opencompass}.

\bibitem[{Ding et~al.(2023)Ding, Chen, Xu, Qin, Zheng, Hu, Liu, Sun, and Zhou}]{ding2023enhancingchatlanguagemodels}
Ning Ding, Yulin Chen, Bokai Xu, Yujia Qin, Zhi Zheng, Shengding Hu, Zhiyuan Liu, Maosong Sun, and Bowen Zhou. 2023.
\newblock \href {https://arxiv.org/abs/2305.14233} {Enhancing chat language models by scaling high-quality instructional conversations}.
\newblock \emph{Preprint}, arXiv:2305.14233.

\bibitem[{Dong et~al.(2024{\natexlab{a}})Dong, Lu, Li, Xia, Yu, Zhou, and Zhou}]{dong2024selfplayexecutionfeedbackimproving}
Guanting Dong, Keming Lu, Chengpeng Li, Tingyu Xia, Bowen Yu, Chang Zhou, and Jingren Zhou. 2024{\natexlab{a}}.
\newblock \href {https://arxiv.org/abs/2406.13542} {Self-play with execution feedback: Improving instruction-following capabilities of large language models}.
\newblock \emph{Preprint}, arXiv:2406.13542.

\bibitem[{Dong et~al.(2024{\natexlab{b}})Dong, Song, Zhu, Qiao, Dou, and Wen}]{dong2024generalinstructionfollowingalignmentretrievalaugmented}
Guanting Dong, Xiaoshuai Song, Yutao Zhu, Runqi Qiao, Zhicheng Dou, and Ji-Rong Wen. 2024{\natexlab{b}}.
\newblock \href {https://arxiv.org/abs/2410.09584} {Toward general instruction-following alignment for retrieval-augmented generation}.
\newblock \emph{Preprint}, arXiv:2410.09584.

\bibitem[{Ge et~al.(2024)Ge, Chan, Wang, Yu, Mi, and Yu}]{ge2024scalingsyntheticdatacreation}
Tao Ge, Xin Chan, Xiaoyang Wang, Dian Yu, Haitao Mi, and Dong Yu. 2024.
\newblock \href {https://arxiv.org/abs/2406.20094} {Scaling synthetic data creation with 1,000,000,000 personas}.
\newblock \emph{Preprint}, arXiv:2406.20094.

\bibitem[{He et~al.(2024{\natexlab{a}})He, Zeng, He, Liang, and Xiao}]{he2024complexsimpleenhancingmulticonstraint}
Qianyu He, Jie Zeng, Qianxi He, Jiaqing Liang, and Yanghua Xiao. 2024{\natexlab{a}}.
\newblock \href {https://arxiv.org/abs/2404.15846} {From complex to simple: Enhancing multi-constraint complex instruction following ability of large language models}.
\newblock \emph{Preprint}, arXiv:2404.15846.

\bibitem[{He et~al.(2024{\natexlab{b}})He, Jin, Wang, Bi, Mandyam, Zhang, Zhu, Li, Xu, Lv, Bhosale, Zhu, Sankararaman, Helenowski, Kambadur, Tayade, Ma, Fang, and Wang}]{he2024multiifbenchmarkingllmsmultiturn}
Yun He, Di~Jin, Chaoqi Wang, Chloe Bi, Karishma Mandyam, Hejia Zhang, Chen Zhu, Ning Li, Tengyu Xu, Hongjiang Lv, Shruti Bhosale, Chenguang Zhu, Karthik~Abinav Sankararaman, Eryk Helenowski, Melanie Kambadur, Aditya Tayade, Hao Ma, Han Fang, and Sinong Wang. 2024{\natexlab{b}}.
\newblock \href {https://arxiv.org/abs/2410.15553} {Multi-if: Benchmarking llms on multi-turn and multilingual instructions following}.
\newblock \emph{Preprint}, arXiv:2410.15553.

\bibitem[{Hendrycks et~al.(2021{\natexlab{a}})Hendrycks, Burns, Basart, Zou, Mazeika, Song, and Steinhardt}]{hendrycks2021measuringmassivemultitasklanguage}
Dan Hendrycks, Collin Burns, Steven Basart, Andy Zou, Mantas Mazeika, Dawn Song, and Jacob Steinhardt. 2021{\natexlab{a}}.
\newblock \href {https://arxiv.org/abs/2009.03300} {Measuring massive multitask language understanding}.
\newblock \emph{Preprint}, arXiv:2009.03300.

\bibitem[{Hendrycks et~al.(2021{\natexlab{b}})Hendrycks, Burns, Kadavath, Arora, Basart, Tang, Song, and Steinhardt}]{hendrycks2021measuringmathematicalproblemsolving}
Dan Hendrycks, Collin Burns, Saurav Kadavath, Akul Arora, Steven Basart, Eric Tang, Dawn Song, and Jacob Steinhardt. 2021{\natexlab{b}}.
\newblock \href {https://arxiv.org/abs/2103.03874} {Measuring mathematical problem solving with the math dataset}.
\newblock \emph{Preprint}, arXiv:2103.03874.

\bibitem[{Hsieh et~al.(2023)Hsieh, Li, Yeh, Nakhost, Fujii, Ratner, Krishna, Lee, and Pfister}]{hsieh2023distillingstepbystepoutperforminglarger}
Cheng-Yu Hsieh, Chun-Liang Li, Chih-Kuan Yeh, Hootan Nakhost, Yasuhisa Fujii, Alexander Ratner, Ranjay Krishna, Chen-Yu Lee, and Tomas Pfister. 2023.
\newblock \href {https://arxiv.org/abs/2305.02301} {Distilling step-by-step! outperforming larger language models with less training data and smaller model sizes}.
\newblock \emph{Preprint}, arXiv:2305.02301.

\bibitem[{Huang et~al.(2025)Huang, Liu, He, Li, Xu, Zhu, Yang, and Zhao}]{huang2025muscimprovingcomplexinstruction}
Hui Huang, Jiaheng Liu, Yancheng He, Shilong Li, Bing Xu, Conghui Zhu, Muyun Yang, and Tiejun Zhao. 2025.
\newblock \href {https://arxiv.org/abs/2502.11541} {Musc: Improving complex instruction following with multi-granularity self-contrastive training}.
\newblock \emph{Preprint}, arXiv:2502.11541.

\bibitem[{Jiang et~al.(2024)Jiang, Wang, Zeng, Zhong, Li, Mi, Shang, Jiang, Liu, and Wang}]{jiang2024followbenchmultilevelfinegrainedconstraints}
Yuxin Jiang, Yufei Wang, Xingshan Zeng, Wanjun Zhong, Liangyou Li, Fei Mi, Lifeng Shang, Xin Jiang, Qun Liu, and Wei Wang. 2024.
\newblock \href {https://arxiv.org/abs/2310.20410} {Followbench: A multi-level fine-grained constraints following benchmark for large language models}.
\newblock \emph{Preprint}, arXiv:2310.20410.

\bibitem[{Kwon et~al.(2023)Kwon, Li, Zhuang, Sheng, Zheng, Yu, Gonzalez, Zhang, and Stoica}]{kwon2023efficient}
Woosuk Kwon, Zhuohan Li, Siyuan Zhuang, Ying Sheng, Lianmin Zheng, Cody~Hao Yu, Joseph~E. Gonzalez, Hao Zhang, and Ion Stoica. 2023.
\newblock Efficient memory management for large language model serving with pagedattention.
\newblock In \emph{Proceedings of the ACM SIGOPS 29th Symposium on Operating Systems Principles}.

\bibitem[{Lambert et~al.(2025)Lambert, Morrison, Pyatkin, Huang, Ivison, Brahman, Miranda, Liu, Dziri, Lyu, Gu, Malik, Graf, Hwang, Yang, Bras, Tafjord, Wilhelm, Soldaini, Smith, Wang, Dasigi, and Hajishirzi}]{lambert2025tulu3pushingfrontiers}
Nathan Lambert, Jacob Morrison, Valentina Pyatkin, Shengyi Huang, Hamish Ivison, Faeze Brahman, Lester James~V. Miranda, Alisa Liu, Nouha Dziri, Shane Lyu, Yuling Gu, Saumya Malik, Victoria Graf, Jena~D. Hwang, Jiangjiang Yang, Ronan~Le Bras, Oyvind Tafjord, Chris Wilhelm, Luca Soldaini, and 4 others. 2025.
\newblock \href {https://arxiv.org/abs/2411.15124} {Tulu 3: Pushing frontiers in open language model post-training}.
\newblock \emph{Preprint}, arXiv:2411.15124.

\bibitem[{Li et~al.(2024{\natexlab{a}})Li, Chiang, Frick, Dunlap, Wu, Zhu, Gonzalez, and Stoica}]{li2024crowdsourceddatahighqualitybenchmarks}
Tianle Li, Wei-Lin Chiang, Evan Frick, Lisa Dunlap, Tianhao Wu, Banghua Zhu, Joseph~E. Gonzalez, and Ion Stoica. 2024{\natexlab{a}}.
\newblock \href {https://arxiv.org/abs/2406.11939} {From crowdsourced data to high-quality benchmarks: Arena-hard and benchbuilder pipeline}.
\newblock \emph{Preprint}, arXiv:2406.11939.

\bibitem[{Li et~al.(2024{\natexlab{b}})Li, Yu, Zhou, Schick, Levy, Zettlemoyer, Weston, and Lewis}]{li2024selfalignmentinstructionbacktranslation}
Xian Li, Ping Yu, Chunting Zhou, Timo Schick, Omer Levy, Luke Zettlemoyer, Jason Weston, and Mike Lewis. 2024{\natexlab{b}}.
\newblock \href {https://arxiv.org/abs/2308.06259} {Self-alignment with instruction backtranslation}.
\newblock \emph{Preprint}, arXiv:2308.06259.

\bibitem[{Lightman et~al.(2023)Lightman, Kosaraju, Burda, Edwards, Baker, Lee, Leike, Schulman, Sutskever, and Cobbe}]{lightman2023letsverifystepstep}
Hunter Lightman, Vineet Kosaraju, Yura Burda, Harri Edwards, Bowen Baker, Teddy Lee, Jan Leike, John Schulman, Ilya Sutskever, and Karl Cobbe. 2023.
\newblock \href {https://arxiv.org/abs/2305.20050} {Let's verify step by step}.
\newblock \emph{Preprint}, arXiv:2305.20050.

\bibitem[{Liu et~al.(2025)Liu, He, Huang, Hu, Liu, Li, Su, and Zheng}]{liu2025aircomplexinstructiongeneration}
Wei Liu, Yancheng He, Hui Huang, Chengwei Hu, Jiaheng Liu, Shilong Li, Wenbo Su, and Bo~Zheng. 2025.
\newblock \href {https://arxiv.org/abs/2502.17787} {Air: Complex instruction generation via automatic iterative refinement}.
\newblock \emph{Preprint}, arXiv:2502.17787.

\bibitem[{Loshchilov and Hutter(2019)}]{loshchilov2019decoupledweightdecayregularization}
Ilya Loshchilov and Frank Hutter. 2019.
\newblock \href {https://arxiv.org/abs/1711.05101} {Decoupled weight decay regularization}.
\newblock \emph{Preprint}, arXiv:1711.05101.

\bibitem[{Meta(2024)}]{grattafiori2024llama3herdmodels}
Meta. 2024.
\newblock \href {https://arxiv.org/abs/2407.21783} {The llama 3 herd of models}.
\newblock \emph{Preprint}, arXiv:2407.21783.

\bibitem[{OpenAI(2024)}]{openai2024gpt4technicalreport}
OpenAI. 2024.
\newblock \href {https://arxiv.org/abs/2303.08774} {Gpt-4 technical report}.
\newblock \emph{Preprint}, arXiv:2303.08774.

\bibitem[{Rein et~al.(2023)Rein, Hou, Stickland, Petty, Pang, Dirani, Michael, and Bowman}]{rein2023gpqagraduatelevelgoogleproofqa}
David Rein, Betty~Li Hou, Asa~Cooper Stickland, Jackson Petty, Richard~Yuanzhe Pang, Julien Dirani, Julian Michael, and Samuel~R. Bowman. 2023.
\newblock \href {https://arxiv.org/abs/2311.12022} {Gpqa: A graduate-level google-proof q\&a benchmark}.
\newblock \emph{Preprint}, arXiv:2311.12022.

\bibitem[{Sun et~al.(2024)Sun, Liu, Li, Wang, Dong, Lin, and Huang}]{sun2024coniferimprovingcomplexconstrained}
Haoran Sun, Lixin Liu, Junjie Li, Fengyu Wang, Baohua Dong, Ran Lin, and Ruohui Huang. 2024.
\newblock \href {https://arxiv.org/abs/2404.02823} {Conifer: Improving complex constrained instruction-following ability of large language models}.
\newblock \emph{Preprint}, arXiv:2404.02823.

\bibitem[{Suzgun et~al.(2022)Suzgun, Scales, Schärli, Gehrmann, Tay, Chung, Chowdhery, Le, Chi, Zhou, and Wei}]{suzgun2022challengingbigbenchtaskschainofthought}
Mirac Suzgun, Nathan Scales, Nathanael Schärli, Sebastian Gehrmann, Yi~Tay, Hyung~Won Chung, Aakanksha Chowdhery, Quoc~V. Le, Ed~H. Chi, Denny Zhou, and Jason Wei. 2022.
\newblock \href {https://arxiv.org/abs/2210.09261} {Challenging big-bench tasks and whether chain-of-thought can solve them}.
\newblock \emph{Preprint}, arXiv:2210.09261.

\bibitem[{Wang et~al.(2024)Wang, Shang, Jain, Wang, Ning, Min, Castelli, Benajiba, and Roth}]{wang2024instructionsconstraintslanguagemodel}
Fei Wang, Chao Shang, Sarthak Jain, Shuai Wang, Qiang Ning, Bonan Min, Vittorio Castelli, Yassine Benajiba, and Dan Roth. 2024.
\newblock \href {https://arxiv.org/abs/2403.06326} {From instructions to constraints: Language model alignment with automatic constraint verification}.
\newblock \emph{Preprint}, arXiv:2403.06326.

\bibitem[{Wang et~al.(2023)Wang, Kordi, Mishra, Liu, Smith, Khashabi, and Hajishirzi}]{wang-etal-2023-self-instruct}
Yizhong Wang, Yeganeh Kordi, Swaroop Mishra, Alisa Liu, Noah~A. Smith, Daniel Khashabi, and Hannaneh Hajishirzi. 2023.
\newblock \href {https://doi.org/10.18653/v1/2023.acl-long.754} {Self-instruct: Aligning language models with self-generated instructions}.
\newblock In \emph{Proceedings of the 61st Annual Meeting of the Association for Computational Linguistics (Volume 1: Long Papers)}, pages 13484--13508, Toronto, Canada. Association for Computational Linguistics.

\bibitem[{White et~al.(2025)White, Dooley, Roberts, Pal, Feuer, Jain, Shwartz-Ziv, Jain, Saifullah, Dey, Shubh-Agrawal, Sandha, Naidu, Hegde, LeCun, Goldstein, Neiswanger, and Goldblum}]{white2025livebenchchallengingcontaminationlimitedllm}
Colin White, Samuel Dooley, Manley Roberts, Arka Pal, Ben Feuer, Siddhartha Jain, Ravid Shwartz-Ziv, Neel Jain, Khalid Saifullah, Sreemanti Dey, Shubh-Agrawal, Sandeep~Singh Sandha, Siddartha Naidu, Chinmay Hegde, Yann LeCun, Tom Goldstein, Willie Neiswanger, and Micah Goldblum. 2025.
\newblock \href {https://arxiv.org/abs/2406.19314} {Livebench: A challenging, contamination-limited llm benchmark}.
\newblock \emph{Preprint}, arXiv:2406.19314.

\bibitem[{Xu et~al.(2023)Xu, Sun, Zheng, Geng, Zhao, Feng, Tao, and Jiang}]{xu2023wizardlmempoweringlargelanguage}
Can Xu, Qingfeng Sun, Kai Zheng, Xiubo Geng, Pu~Zhao, Jiazhan Feng, Chongyang Tao, and Daxin Jiang. 2023.
\newblock \href {https://arxiv.org/abs/2304.12244} {Wizardlm: Empowering large language models to follow complex instructions}.
\newblock \emph{Preprint}, arXiv:2304.12244.

\bibitem[{Xu et~al.(2024)Xu, Jiang, Niu, Deng, Poovendran, Choi, and Lin}]{xu2024magpie}
Zhangchen Xu, Fengqing Jiang, Luyao Niu, Yuntian Deng, Radha Poovendran, Yejin Choi, and Bill~Yuchen Lin. 2024.
\newblock Magpie: Alignment data synthesis from scratch by prompting aligned llms with nothing.
\newblock \emph{arXiv preprint arXiv:2406.08464}.

\bibitem[{Yang et~al.(2025{\natexlab{a}})Yang, Li, Yang, Zhang, Hui, Zheng, Yu, Gao, Huang, Lv, Zheng, Liu, Zhou, Huang, Hu, Ge, Wei, Lin, Tang, Yang, Tu, Zhang, Yang, Yang, Zhou, Zhou, Lin, Dang, Bao, Yang, Yu, Deng, Li, Xue, Li, Zhang, Wang, Zhu, Men, Gao, Liu, Luo, Li, Tang, Yin, Ren, Wang, Zhang, Ren, Fan, Su, Zhang, Zhang, Wan, Liu, Wang, Cui, Zhang, Zhou, and Qiu}]{yang2025qwen3technicalreport}
An~Yang, Anfeng Li, Baosong Yang, Beichen Zhang, Binyuan Hui, Bo~Zheng, Bowen Yu, Chang Gao, Chengen Huang, Chenxu Lv, Chujie Zheng, Dayiheng Liu, Fan Zhou, Fei Huang, Feng Hu, Hao Ge, Haoran Wei, Huan Lin, Jialong Tang, and 41 others. 2025{\natexlab{a}}.
\newblock \href {https://arxiv.org/abs/2505.09388} {Qwen3 technical report}.
\newblock \emph{Preprint}, arXiv:2505.09388.

\bibitem[{Yang et~al.(2025{\natexlab{b}})Yang, Wan, Zhong, Huang, Liang, and Quan}]{yang2025fusechat30preferenceoptimizationmeets}
Ziyi Yang, Fanqi Wan, Longguang Zhong, Canbin Huang, Guosheng Liang, and Xiaojun Quan. 2025{\natexlab{b}}.
\newblock \href {https://arxiv.org/abs/2503.04222} {Fusechat-3.0: Preference optimization meets heterogeneous model fusion}.
\newblock \emph{Preprint}, arXiv:2503.04222.

\bibitem[{Zellers et~al.(2019)Zellers, Holtzman, Bisk, Farhadi, and Choi}]{zellers2019hellaswagmachinereallyfinish}
Rowan Zellers, Ari Holtzman, Yonatan Bisk, Ali Farhadi, and Yejin Choi. 2019.
\newblock \href {https://arxiv.org/abs/1905.07830} {Hellaswag: Can a machine really finish your sentence?}
\newblock \emph{Preprint}, arXiv:1905.07830.

\bibitem[{Zhang et~al.(2024)Zhang, Yu, Fu, Huang, and Li}]{zhang2024iopoempoweringllmscomplex}
Xinghua Zhang, Haiyang Yu, Cheng Fu, Fei Huang, and Yongbin Li. 2024.
\newblock \href {https://arxiv.org/abs/2411.06208} {Iopo: Empowering llms with complex instruction following via input-output preference optimization}.
\newblock \emph{Preprint}, arXiv:2411.06208.

\bibitem[{Zheng et~al.(2024)Zheng, Zhang, Zhang, Ye, Luo, Feng, and Ma}]{zheng2024llamafactory}
Yaowei Zheng, Richong Zhang, Junhao Zhang, Yanhan Ye, Zheyan Luo, Zhangchi Feng, and Yongqiang Ma. 2024.
\newblock \href {http://arxiv.org/abs/2403.13372} {Llamafactory: Unified efficient fine-tuning of 100+ language models}.
\newblock In \emph{Proceedings of the 62nd Annual Meeting of the Association for Computational Linguistics (Volume 3: System Demonstrations)}, Bangkok, Thailand. Association for Computational Linguistics.

\bibitem[{Zhou et~al.(2023)Zhou, Lu, Mishra, Brahma, Basu, Luan, Zhou, and Hou}]{zhou2023instructionfollowingevaluationlargelanguage}
Jeffrey Zhou, Tianjian Lu, Swaroop Mishra, Siddhartha Brahma, Sujoy Basu, Yi~Luan, Denny Zhou, and Le~Hou. 2023.
\newblock \href {https://arxiv.org/abs/2311.07911} {Instruction-following evaluation for large language models}.
\newblock \emph{Preprint}, arXiv:2311.07911.

\end{thebibliography}

\appendix

\section*{Appendix}
\label{sec:appendix}

\section{Pipeline of DecIF}
As shown in Algorithm \ref{alg::DecIF}, we formally present the pipeline of DecIF. 

\section{Detailed Description of Baselines}
\label{sec::baseline}
In this paper, we compare DecIF with a comprehensive set of baselines to thoroughly validate its effectiveness. Specifically, 

\textbf{ShareGPT}\citep{vicuna2023} is an open-source and multi-turn conversation dataset that contains 52k user-shared chatting histories with GPT-4. We randomly select 10k subset to serve as the baseline.

\textbf{Evol-Instruct} \citep{xu2023wizardlmempoweringlargelanguage} is a large-scale complex instruction dataset that evolves existing instructions across both depth and breadth. We also randomly select a 10k subset to serve as the baseline.

\textbf{Conifer} \citep{sun2024coniferimprovingcomplexconstrained} is a 13k instruction-following dataset synthesized from seed instructions derived from ShareGPT, using a three-stage process involving query reframing, constraint generation, and recombination. We directly utilize the full set as the baseline.

\textbf{AIR} \citep{liu2025aircomplexinstructiongeneration} collects initial instructions from existing documents and performs iterative instruction refinement to construct a 10k instruction-following dataset. We use the full set of the data as the baseline.

\textbf{Condor} \citep{cao2025condorenhancellmalignment} incorporate world knowledge tree to synthesize a large-scale instruction data. We directly use the open-source 20k data to serve as the baseline.

\textbf{AutoIF} \citep{dong2024selfplayexecutionfeedbackimproving} manually design a variety of constraints and innovatively employ Python functions to evaluate the responses. In this paper, we directly report the performance in \citep{an2025ultraifadvancinginstructionfollowing}.

\textbf{UltraIF} \citep{an2025ultraifadvancinginstructionfollowing} incorporate existing documents and datasets to train the UltraComposer and synthesize a large-scale instruction-following data. We also directly report the performance in its original paper.

For more challenging comparison and validate the generalizability of DecIF, we also utilize \textbf{Tulu-3} \citep{lambert2025tulu3pushingfrontiers} and \textbf{Magpie} \citep{xu2024magpie} as our baselines.

\begin{algorithm*}[htbp]
\centering
\begin{algorithmic}[1]
    \REQUIRE 
        LLM $\theta$, constraint pool $\mathcal{C}$, iterations $T$, \\
        meta-domains per iter. $D$, meta-requests per domain $R$, \\
        meta-scenarios per request $S$, constraint dist. $\vec{p} = [p_1, p_2, p_3, p_4, p_5]$

    \ENSURE Generated dataset $\mathcal{D}$

    \STATE Initialize: $\mathcal{M}_d, \mathcal{M}_r, \mathcal{M}_s, \mathcal{I}, \mathcal{D} \gets \emptyset$, $i \gets 0$
    
    \WHILE{$i < T$}
        \STATE $\mathcal{D}_{\text{new}} \sim \theta_{\text{generate}}(\cdot \mid D)$
        \STATE $\mathcal{M}_d \gets \mathcal{M}_d \cup \mathcal{D}_{\text{new}}$, $i \gets i + 1$
    \ENDWHILE

    \FORALL{$d \in \mathcal{M}_d$}
        \STATE $\mathcal{R}_{\text{new}} \sim \theta_{\text{generate}}(\cdot \mid d, R)$
        \STATE $\mathcal{M}_r \gets \mathcal{M}_r \cup \mathcal{R}_{\text{new}}$
    \ENDFOR

    \FORALL{$r \in \mathcal{M}_r$}
        \STATE $\mathcal{S}_{\text{new}} \sim \theta_{\text{generate}}(\cdot \mid r, S)$
        \STATE $\mathcal{M}_s \gets \mathcal{M}_s \cup \mathcal{S}_{\text{new}}$
    \ENDFOR

    \FORALL{$s \in \mathcal{M}_s$}
        \STATE Sample $k \in \{1,\dots,5\}$ with $\vec{p}$
        \STATE Select $\mathcal{C}_k \subseteq \mathcal{C}$ with $|\mathcal{C}_k| = k$
        \STATE $i_0 \sim \theta_{\text{generate}}(s, \mathcal{C}_k)$
        \WHILE{$\theta_{\text{conflict\_detect}}(i_0, \mathcal{C}_k) = \text{True}$}
            \STATE $i_0 \sim \theta_{\text{refine}}(i_0)$
        \ENDWHILE
        \STATE $\mathcal{I} \gets \mathcal{I} \cup \{i_0\}$
    \ENDFOR

    \FORALL{$i \in \mathcal{I}$}
        \STATE $y \sim \theta_{\text{generate}}(i)$, $Q \sim \theta_{\text{decompose}}(i)$
        \STATE $\mathcal{E} \sim \theta_{\text{evaluate}}(Q, y)$
        \IF{$\exists q \in Q : \mathcal{E}(q) = \text{NO}$}
            \STATE skip $(i, y)$
        \ELSE
            \STATE $\mathcal{D} \gets \mathcal{D} \cup \{(i, y)\}$
        \ENDIF
    \ENDFOR

    \RETURN Final dataset $\mathcal{D}$
\end{algorithmic}
\caption{The workflow of DecIF}
\label{alg::DecIF}
\end{algorithm*}

\begin{table*}[htbp]
\centering
\resizebox{\linewidth}{!}{%
\begin{tabular}{@{}lccccccccccc@{}}
\toprule
                                  &                                   & \multicolumn{4}{c}{\textbf{IFEval}} & \multicolumn{3}{c}{\textbf{Multi-IF}} & \multicolumn{2}{c}{\textbf{FollowBench}} & \textbf{LiveBench} \\ \cmidrule(l){3-12} 
\multirow{-2}{*}{\textbf{Method}} & \multirow{-2}{*}{\textbf{\#Data}} & Pr (S)  & In (S)  & Pr (L) & In (L) & Turn 1      & Turn 2     & Turn 3     & HSR                 & SSR                & Score              \\ \midrule
\multicolumn{12}{l}{\cellcolor[HTML]{EFEFEF}\textit{Results of Tulu-3 instruction-following subset}} \\
Qwen-3-4B               & 30k                               & 75.42   & 82.37   & \textbf{78.37}  & \textbf{85.01}   & \textbf{78.80}       & 60.65      &  44.54     & 61.96               & 72.41              & \textbf{53.10}              \\
Qwen-3-14B & 30k & 77.26 & 83.93 & 81.52 & \textbf{87.92} & 67.71 & 61.47 & 49.86 & \textbf{70.72} & \textbf{79.62} & 44.90 \\
Gemma-3-4B               & 30k                               & 60.26   & 69.90   & 63.96  & 73.74  & 67.12       & 41.92      & 30.25      & 50.85               & 60.86              & \textbf{48.20}              \\
Gemma-3-12B               & 30k                               & 68.95   & 77.58   & 72.27  & 80.34  & 75.85       & 51.54      & 39.89      & 64.43               & \textbf{75.82}              & 59.60              \\ \midrule
\multicolumn{12}{l}{\cellcolor[HTML]{EFEFEF}\textit{Results of DecIF instruction-following data}} \\
Qwen-3-4B               & 30k                               & \textbf{76.16}   & \textbf{82.85}   & 77.82  & 84.41  & 78.21       & \textbf{64.73}      & \textbf{51.74}      & \textbf{62.50}               & \textbf{72.85}              & 50.70              \\
Qwen-3-14B & 30k & \textbf{80.04} & \textbf{85.49} & \textbf{82.62} & 87.77 & \textbf{83.00} & \textbf{71.09} & \textbf{60.27} & 69.43 & 78.74 & \textbf{61.10} \\
Gemma-3-4B               & 30k                               & \textbf{65.62}   & \textbf{74.10}   & \textbf{68.76}  & \textbf{76.98}  & \textbf{70.58}       & \textbf{48.03}      & \textbf{35.63}      & \textbf{52.27}               & \textbf{61.05}              & 47.30              \\
Gemma-3-12B               & 30k                               & \textbf{72.27}   & \textbf{79.74}   & \textbf{74.86}  & \textbf{82.13}  & \textbf{76.32}       & \textbf{55.17}      & \textbf{41.63}      & \textbf{65.01}               & 75.37              & \textbf{61.20}              \\ \bottomrule
\end{tabular}
}
\caption{The results on four instruction-following benchmarks under Qwen-3 and Gemma-3 series models. Note that we utilize LLaMA-3.1-70B-Instruct as supervised model.}
\label{tab::more_base_model}
\end{table*}

\section{Detailed Description of Evaluation Benchmarks}
\label{sec::benchmark}
We utilize OpenCompass \citep{2023opencompass} to evaluate most of the benchmarks.

For the evaluation of instruction-following capability, we utilize the following benchmarks:

\textbf{IFEval} \citep{zhou2023instructionfollowingevaluationlargelanguage} is an easily producible benchmark specifically designed to assess the instruction-following capabilities of LLMs. It consists of approximately 500 prompts, each containing 25 types of verifiable instructions. In our evaluation, we employ both loose and strict accuracy metrics at both the prompt and instruction levels.

\textbf{Multi-IF} \citep{he2024multiifbenchmarkingllmsmultiturn} is a benchmark designed to evaluate LLMs' proficiency in following multi-turn and multilingual instructions. It comprises 4,501 multilingual conversations, each consisting of three turns. We report the average accuracy across all languages for each of the three rounds in the experiment.

\textbf{FollowBench} \citep{jiang2024followbenchmultilevelfinegrainedconstraints} is a multi-level, fine-grained constraint-following benchmark designed for LLMs. It integrates five distinct types of fine-grained constraints, Content, Situation, Style, Format, and Example, and emphasizes a multi-level mechanism in the construction of instruction prompts. In our experiments, we utilize GPT-4o-2025-03-26 to evaluate whether the outputs of LLMs satisfy each individual constraint.

\textbf{LiveBench} \citep{white2025livebenchchallengingcontaminationlimitedllm} is an LLM benchmark that encompasses a wide array of challenging tasks, including math, coding, reasoning, language, instruction-following, and data analysis, with answers automatically scored based on objective ground-truth values. We use the instruction-following subset to asses the instruction-following capability and utilize all data to evaluate the general capability.

For other common abilities, we incorporate the following benchmarks to broadly assess the models:

\textbf{GSM8K} \citep{cobbe2021trainingverifierssolvemath} consists of 8.5K high-quality, multilingual grade school math word problems, meticulously crafted to evaluate the multi-step mathematical reasoning proficiency of language models. In the experiment, we report the overall accuracy achieved by the models.

\textbf{MATH} \citep{hendrycks2021measuringmathematicalproblemsolving} is a challenging benchmark containing 12,500 high school-level mathematics problems spanning algebra, geometry, and more. Each problem includes a textual description and a step-by-step solution. The benchmark is designed to assess the ability of language models to perform formal mathematical problem-solving. In the experiment, we report the accuracy of final answers predicted by the models.

\textbf{HumanEval} \citep{chen2021evaluatinglargelanguagemodels} comprises 164 programming problems, each including function signatures, docstrings, bodies, and unit tests, with an average of 7.7 tests per problem. It is designed to evaluate the coding capabilities of LLMs. HumanEval assesses the models' ability to synthesize programs from docstrings, testing their language comprehension, reasoning, algorithmic thinking, and elementary mathematics skills. In the experiment, we report the Pass@1 score on HumanEval.

\begin{table*}[htbp]
\centering
\resizebox{\linewidth}{!}{%
\begin{tabular}{@{}lccccccccccc@{}}
\toprule
                                  &                                   & \multicolumn{4}{c}{\textbf{IFEval}} & \multicolumn{3}{c}{\textbf{Multi-IF}} & \multicolumn{2}{c}{\textbf{FollowBench}} & \textbf{LiveBench} \\ \cmidrule(l){3-12} 
\multirow{-2}{*}{\textbf{Method}} & \multirow{-2}{*}{\textbf{\#Data}} & Pr (S)  & In (S)  & Pr (L) & In (L) & Turn 1      & Turn 2     & Turn 3     & HSR                 & SSR                & Score              \\ \midrule
\multicolumn{12}{l}{\cellcolor[HTML]{EFEFEF}\textit{Results of Qwen-3-32B}} \\
Qwen-3-32B (w/o think) & 30k & \textbf{84.84} & \textbf{89.57} & 87.62 & 91.61 & 86.95 & 78.79 & 71.64 & \textbf{76.01} & \textbf{82.37} & 83.40 \\
Qwen-3-32B (w/ think) & 30k & 84.29 & 88.85 & \textbf{88.91} & \textbf{92.33} & \textbf{87.63} & \textbf{79.62} & \textbf{71.98} & 75.52 & 80.58 & \textbf{83.90} \\ \midrule
\multicolumn{12}{l}{\cellcolor[HTML]{EFEFEF}\textit{Results of non-thinking instruction-following data}} \\
Qwen-3-4B & 30k & 74.31 & \textbf{82.13} & 78.19 & \textbf{85.01} & 74.16 & 61.43 & \textbf{50.71} & 65.84 & 75.27 & 51.90 \\
Qwen-3-8B               & 30k & \textbf{78.56} & 84.65 & 81.89 & \textbf{87.53} & 80.86 & \textbf{66.39} & \textbf{55.32} & \textbf{72.95} & \textbf{80.24} & 61.20               \\
Qwen-3-14B & 30k & \textbf{80.41} & 85.61 & \textbf{84.66} & \textbf{88.97} & 82.02 & 71.71 & \textbf{62.21} & \textbf{73.58} & \textbf{81.14} & 67.40 \\ \midrule
\multicolumn{12}{l}{\cellcolor[HTML]{EFEFEF}\textit{Results of long-CoT instruction-following data}} \\
Qwen-3-4B & 30k & \textbf{74.86} & 82.01 & 78.19 & 84.53 & \textbf{76.32} & \textbf{62.00} & 46.71 & \textbf{66.56} & 74.46 & \textbf{61.70} \\
Qwen-3-8B               & 30k & 78.19 & \textbf{84.77} & 81.89 & 87.29 & \textbf{81.91} & 63.50 & 49.66 & 70.84 & 77.51 & \textbf{68.20}             \\
Qwen-3-14B & 30k & 79.48 & 85.61 & 82.62 & 87.89 & \textbf{84.51} & \textbf{73.04} & 54.41 & 71.96 & 78.64 & \textbf{71.90} \\ \bottomrule
\end{tabular}
}
\caption{The results on four instruction-following benchmarks under various base models with long-CoT data. Note that we utilize Qwen-3-32B as supervised model.}
\label{tab::thinking_result}
\end{table*}

\textbf{BBH} \citep{suzgun2022challengingbigbenchtaskschainofthought} is a clean, challenging, and tractable subset benchmark filtered from Big Bench, comprising 23 types of difficult tasks and a total of 6,511 evaluation examples. BBH primarily focuses on comprehensively assessing the reasoning capabilities and problem-solving skills of models. In the experiment, we report accuracy metrics on BBH.

\textbf{HellaSwag} \citep{zellers2019hellaswagmachinereallyfinish} is a challenging benchmark designed to evaluate the commonsense reasoning capabilities of language models through sentence completion tasks. It comprises approximately 70,000 multiple-choice questions, each presenting a context followed by four possible endings. Only one of these endings is correct. In the experiments, we report the model's accuracy in selecting the correct ending.

\textbf{GPQA} \citep{rein2023gpqagraduatelevelgoogleproofqa} is a highly challenging subset of the GPQA benchmark, consisting of 198 multiple-choice questions across biology, physics, and chemistry. The Diamond subset is distinguished by its stringent validation criteria. In the experiments, we report the model's accuracy in selecting the correct answer.

\textbf{MMLU} \citep{hendrycks2021measuringmassivemultitasklanguage} is a comprehensive benchmark designed to evaluate the multitask accuracy of language models across a diverse set of academic and professional subjects. It encompasses 57 tasks, including areas such as elementary mathematics, U.S. history, computer science, law, and medicine, totaling approximately 15,908 multiple-choice questions. In the experiments, we report the average accuracy across all tasks.

\textbf{Arena Hard} \citep{li2024crowdsourceddatahighqualitybenchmarks} is an automated LLM benchmark comprising 500 challenging user queries, carefully curated to evaluate the comprehensive performance of LLMs in user dialogue scenarios. In the experiment, we use GPT-4o-2025-03-26 as the judge model and report the win rate of our models compared to the baseline model (GPT-4-0314).

\section{Further Experiments}
\label{sec::further_experiment}
\subsection{More Results on Various Base Models}
To further validate the effectiveness of DecIF, we conduct experiments across a variety of base models. Specifically, we compare DecIF with the instruction-following subset of Tulu-3, using several state-of-the-art models such as Qwen-3-4B, Qwen-3-14B, Gemma-3-4B, and Gemma-3-12B. Detailed results are presented in Table \ref{tab::more_base_model}. DecIF consistently achieves strong performance across most benchmarks, demonstrating its broad applicability and effectiveness. 

\subsection{Long-CoT Training}
Recently, the long-chain-of-thought (long-CoT) training approach has demonstrated strong performance across various domains, particularly in mathematical reasoning and code generation. In this section, we investigate its potential within instruction-following scenarios. Specifically, we employ Qwen-3-32B to generate long-CoT responses for our synthesized instructions and use these responses to train a range of base models.
As presented in Table \ref{tab::thinking_result}, models trained with long-CoT data exhibit a performance drop on IFEval compared to non-thinking baselines, yet achieve substantial improvements on LiveBench. Furthermore, in the Multi-IF benchmark, thinking models show superior performance in the first turn, but their effectiveness diminishes as the number of interaction turns increases. In contrast, non-thinking models perform relatively better in later turns. We hypothesize that this is due to the single-turn nature of our training data, thinking models are not exposed to multi-turn long-CoT dialogues during training, which limits their ability to maintain coherent reasoning over extended interactions.
Overall, incorporating long-CoT data during the SFT stage yields notable gains for certain input types, with minimal adverse effects on others. We believe that long-CoT data holds significant promise for advancing instruction-following capabilities and warrants further exploration as a direction for future research.

\section{Experimental Details}

\subsection{Implementation Details of DecIF}
\label{sec::inplementation}
In our instruction data synthesis process, we utilize vLLM \citep{kwon2023efficient} for efficient inference. We prompt LLMs to generate 25 domains per iteration over 1000 iterations, followed by deduplication of any repeated domains. For different LLMs, this process typically results in the synthesis of 140 to 170 distinct real-life domains. Subsequently, we prompt LLMs to generate approximately 30 meta-requests for each domain. For each meta-request, we then instruct LLMs to produce 20 distinct meta-scenarios. Ultimately, different supervised models yield approximately 10k requests, each with 20 distinct scenarios. If further scaling up is required, it is possible to flexibly prompt LLMs to generate a greater quantity of content. When synthesizing the final instructions, we randomly sample 1 to 5 constraints of different types from the constraint pools. The proportion of instructions with 1 to 5 constraints is set at 0.2, 0.3, 0.3, 0.1, and 0.1, respectively. For the entire data synthesis process, we use a temperature of 0.6, top\_p of 0.95, and max\_tokens of 4096. As for the model evaluation process, we consistently employ greedy decoding to ensure stable and reliable assessments.

\subsection{Training Details}
\label{sec::training}
For the training of instruction-following data, we follow \citep{an2025ultraifadvancinginstructionfollowing} and perform full fine-tuning with a learning rate of 1e-5. The maximum token length is set to 8192. We use AdamW \citep{loshchilov2019decoupledweightdecayregularization} as the optimizer with a warmup ratio of 0.03 and train for 3 epochs. Additionally, we employ a LinearLR scheduler at the beginning, transitioning to CosineAnnealingLR towards the end of training. For the training of large-scale general-purpose data, we follow \citep{lambert2025tulu3pushingfrontiers} and perform full fine-tuning with a learning rate of 5e-6. The maximum token length is set to 4096. We use AdamW as the optimizer with a warmup ratio of 0.03 and train for 2 epochs. Additionally, we employ a LinearLR scheduler throughout the entire training process. We utilize LLaMA-Factory \citep{zheng2024llamafactory} framework for all the training process.

\begin{table*}[htbp]
\centering
\renewcommand{\arraystretch}{1.1}
\setlength{\tabcolsep}{5pt}
\begin{tabular}{@{}p{3.8cm}p{9.2cm}@{}}
\toprule
\textbf{Constraint Type} & \textbf{Description} \\
\midrule
Include Keywords & Include specific keyword(s) in the response. \\
\addlinespace[3pt]
Keyword Frequency & Keywords should appear specific times. \\
\addlinespace[3pt]
Forbidden Words & Exclude specified keyword(s). \\
\addlinespace[3pt]
Letter Frequency & Specific letter should appear specific times. \\
\addlinespace[3pt]
Response Language & Response must be in specified language only. \\
\addlinespace[3pt]
Number Paragraphs & Contain specific paragraphs separated by symbol. \\
\addlinespace[3pt]
Number Words & Response length: at least/around/at most X words. \\
\addlinespace[3pt]
Number Sentences & Response length: at least/around/at most X sentences. \\
\addlinespace[3pt]
Mixed & Specific paragraphs starting with required phrases. \\
\addlinespace[3pt]
Postscript & Must include specified postscript markers. \\
\addlinespace[3pt]
Number Placeholder & Must contain [specified] placeholders. \\
\addlinespace[3pt]
Number Bullets & Must contain specified bullet points. \\
\addlinespace[3pt]
Title & Must include title in specified format. \\
\addlinespace[3pt]
Choose From & Response must be one of specified options. \\
\bottomrule
\end{tabular}
\caption{Response Constraint Types (Part 1 of 2)}
\label{tab::response_constraints_1}
\end{table*}

\begin{table*}[htbp]
\centering
\renewcommand{\arraystretch}{1.1}
\setlength{\tabcolsep}{5pt}
\begin{tabular}{@{}p{3.8cm}p{9.2cm}@{}}
\toprule
\textbf{Constraint Type} & \textbf{Description} \\
\midrule
Highlighted Section & Highlight sections with specified format. \\
\addlinespace[3pt]
Multiple Sections & Sections marked with specified splitters. \\
\addlinespace[3pt]
Multiple Format & Response format: JSON/Table/HTML/XML/LaTeX/Markdown. \\
\addlinespace[3pt]
Repeat Prompt & First repeat request verbatim, then answer. \\
\addlinespace[3pt]
Two Responses & Provide two responses separated by symbol. \\
\addlinespace[3pt]
All Uppercase & Response in CAPITAL LETTERS only. \\
\addlinespace[3pt]
All Lowercase & Response in lowercase only. \\
\addlinespace[3pt]
Allcapital Words & Words in ALL CAPS appear X times. \\
\addlinespace[3pt]
End Checker & Response must end with specified phrase. \\
\addlinespace[3pt]
Start Checker & Response must start with specified phrase. \\
\addlinespace[3pt]
Quotation & Response wrapped in specified marks. \\
\addlinespace[3pt]
No Commas & Response must not contain commas. \\
\addlinespace[3pt]
Role-based & Simulate character traits/behaviors. \\
\addlinespace[3pt]
Scenario-based & Response for specific situation. \\
\addlinespace[3pt]
Style & Response in specified style/tone. \\
\addlinespace[3pt]
Audience & Response tailored to specific audience. \\
\bottomrule
\end{tabular}
\caption{Response Constraint Types (Part 2 of 2)}
\label{tab::response_constraints_2}
\end{table*}
\subsection{Constraint Types of DecIF}
\label{sec::contraint_type}
As shown in Table \ref{tab::response_constraints_1} and \ref{tab::response_constraints_2}, we employ response constraint types which are primarily derived from \citep{zhou2023instructionfollowingevaluationlargelanguage}, to construct instruction-following data.

\subsection{Prompt Templates of DecIF}
\label{sec::prompt_template}
We utilize the following prompt templates to support DecIF to construct high-quality instruction data without relying on any external resources.

\begin{figure*}[t] 
\begin{tcolorbox}[
    colback=white, 
    colframe=gray!50, 
    arc=4pt, 
    boxrule=1pt, 
    left=6pt, right=6pt, 
    title={\bfseries Prompt Template for Meta-Domains Generation}, 
    fonttitle=\sffamily,
    width=\textwidth 
]
Generate exactly \{number of domains\} real‑world domains that these tasks might address.

Criteria:

- Cover everyday life comprehensively.

- Each domain is broad, distinct, and has clear practical value.

- All tasks within the domain must be solvable by a language model.

Examples:

- Education

- Healthcare

- Finance

- Technology

- Travel 

- Computer Science

- Artificial Intelligence

- Data Science

- Mathematics

*Strict Output Format Requirements*

- Each domain must start with a hyphen followed by a space ("- ")

- Do not number the items

- Do not include any additional text, explanations, or formatting

- Do not repeat any examples from the input

- Maintain exactly one domain per line

Output exactly \{number of domains\} domains in this format:

- Domain A

- Domain B

- Domain C

...
\end{tcolorbox}
\end{figure*}

\begin{figure*}[t] 
\begin{tcolorbox}[
    colback=white, 
    colframe=gray!50, 
    arc=4pt, 
    boxrule=1pt, 
    left=6pt, right=6pt, 
    title={\bfseries Prompt Template for Meta-Requests Generation}, 
    fonttitle=\sffamily,
    width=\textwidth 
]
Generate nearly \{number of requests\} diverse short task instructions (meta requests) specifically for the \{domain\} domain.

*Requirements* 

1. Each instruction must be **less than 4 words**, specific, unique, realistic, common, and *model‑solvable*.

2. All instructions must be relevant to the \{domain\} domain.

3. No duplicate instructions within the output.

4. Instructions should be clear and actionable (avoid vague commands like "Do task").

*Example output for "Education" domain (Strictly follow this format and use lowercase letters)*  

- explain the math concept

- grade student essays

- create lesson plan

- suggest teaching methods  

- recommend educational apps 

Now generate nearly \{number of requests\} diverse meta requests specifically for the \{domain\} domain:
\end{tcolorbox}
\end{figure*}

\begin{figure*}[t] 
\begin{tcolorbox}[
    colback=white, 
    colframe=gray!50, 
    arc=4pt, 
    boxrule=1pt, 
    left=6pt, right=6pt, 
    title={\bfseries Prompt Template for Meta-Scenarios Generation}, 
    fonttitle=\sffamily,
    width=\textwidth 
]
For the meta request \{meta request\}, generate \{number of scenarios\} diverse and realistic scenarios that would require this action in everyday life. Each scenario should:

1. Be specific with clear context (who, what, where, why)

2. Be from different domains (work, education, personal life, etc.)

3. Be 1-2 sentences maximum

4. Use hyphen formatting (- ...) for each scenario

An Example:

Meta request: create guide

- A fitness trainer needs to create a workout guide for elderly clients at a local community center

A software company wants to create an onboarding guide for new remote employees

A parent needs to create a morning routine guide for their children before school

...

Now generate scenarios for:

Meta request: \{meta request\}
\end{tcolorbox}
\end{figure*}

\begin{figure*}[t] 
\begin{tcolorbox}[
    colback=white, 
    colframe=gray!50, 
    arc=4pt, 
    boxrule=1pt, 
    left=6pt, right=6pt, 
    title={\bfseries Prompt Template for Instructions Generation}, 
    fonttitle=\sffamily,
    width=\textwidth 
]
Create a verifiable instruction that the following persona might ask you to do:

\{meta scenarios\}

An example:

Write down the names of two famous international badminton mixed doubles tournaments and your answer should be all in capital words.

Note:

1. The above example is not tied to any particular persona, but you should create one that is unique and specific to the given persona.

2. The instruction should contain all the following verifiable constraint(s):

\{the selected constraint(s)\}

3. Your output should start with "User instruction:". Your output should not include an answer to the instruction.
\end{tcolorbox}
\end{figure*}

\begin{figure*}[t] 
\begin{tcolorbox}[
    colback=white, 
    colframe=gray!50, 
    arc=4pt, 
    boxrule=1pt, 
    left=6pt, right=6pt, 
    title={\bfseries Prompt Template for Consistency Judgement}, 
    fonttitle=\sffamily,
    width=\textwidth 
]
You are an expert in analyzing instructions for internal conflicts. Your task is to analyze the following instruction:

\{instruction\}

Follow these steps:

1. Check if there are any conflicting requirements (e.g., requiring both Chinese and English).

2. If there is a conflict, refine the instruction to resolve it. The refined instruction must be clear, concise, and free of any explanatory text.

3. If there is no conflict, return the original instruction unchanged.

4. Format your response as follows:

- Original: <original\_instruction>

- Conflict: True/False

- Refined: <refined\_instruction>

Ensure that the 'Refined' field contains ONLY the refined instruction without any additional explanations or context.
\end{tcolorbox}
\end{figure*}

\begin{figure*}[t] 
\begin{tcolorbox}[
    colback=white, 
    colframe=gray!50, 
    arc=4pt, 
    boxrule=1pt, 
    left=6pt, right=6pt, 
    title={\bfseries Prompt Template for Instruction-Following Data Responses Generation}, 
    fonttitle=\sffamily,
    width=\textwidth 
]
You are an expert tasked with answering the given query. Please provide a clear and concise response directly, without introductory phrases such as 'What a great question', 'Here is the answer,' or similar expressions. 

Focus solely on addressing the query.

Now please answer the given query while strictly following its inside constraints.

[Query] \{instruction\}
\end{tcolorbox}
\end{figure*}

\begin{figure*}[t] 
\begin{tcolorbox}[
    colback=white, 
    colframe=gray!50, 
    arc=4pt, 
    boxrule=1pt, 
    left=6pt, right=6pt, 
    title={\bfseries Prompt Template for General-Purpose Data Responses Generation}, 
    fonttitle=\sffamily,
    width=\textwidth 
]
You are an expert tasked with answering the given query.

Please provide a clear and accurate response to the given query.

[Query] \{instruction\}
\end{tcolorbox}
\end{figure*}

\begin{figure*}[t] 
\begin{tcolorbox}[
    colback=white, 
    colframe=gray!50, 
    arc=4pt, 
    boxrule=1pt, 
    left=6pt, right=6pt, 
    title={\bfseries Prompt Template for Instructions Decomposition}, 
    fonttitle=\sffamily,
    width=\textwidth 
]
You are now an Evaluation Criteria Designer, tasked with breaking down complex instructions into granular evaluation questions. These questions will be used to assess whether a response meets the requirements of the given instruction.

Your task is as follows:

Analyze the provided instruction and identify all atomic-level requirements including: content specifications, formatting constraints, stylistic guidelines, factual accuracy checks, logical consistency requirements, and any other explicit or implicit conditions that must be satisfied.

For each requirement or constraint, create a clear, concise evaluation question that can be answered with a simple "yes" or "no."

Ensure the questions are specific, actionable, and free of any explanations or additional context.

Instruction:

\{instruction\}

Output Format:

Each evaluation question should be on a new line.
Questions must be phrased in a way that allows for a binary ("yes" or "no") answer.

Example Output:

Does the response include at least three sources?

Is the response under 100 words?

Now, proceed with the breakdown.
\end{tcolorbox}
\end{figure*}

\begin{figure*}[t] 
\begin{tcolorbox}[
    colback=white, 
    colframe=gray!50, 
    arc=4pt, 
    boxrule=1pt, 
    left=6pt, right=6pt, 
    title={\bfseries Prompt Template for Responses Judgement}, 
    fonttitle=\sffamily,
    width=\textwidth 
]
You are a Quality Assurance Specialist for AI responses. Your task is to rigorously evaluate whether a response meets all specified criteria. You must be thorough and impartial in your assessments.

Evaluation Guidelines:

1. Examine each criterion independently

2. Be strict but fair - only mark 'YES' if the response fully satisfies the criterion

3. Ignore any stylistic preferences not explicitly listed in the criteria

4. Focus exclusively on the criteria provided

Instruction:

\{instruction\}

Response to evaluate:

\{response\}

Evaluation criteria:

\{criteria\}

Your task:

For each criterion above, output ONLY either 'YES' or 'NO' on its own line, in order.

Begin evaluation:
\end{tcolorbox}
\end{figure*}

\end{document}